\documentclass{article}

\PassOptionsToPackage{table}{xcolor}
\usepackage{etoolbox}       %
\usepackage[utf8]{inputenc} 

\usepackage{graphicx}  %

\usepackage[parfill]{parskip}
\usepackage{amsmath,amsthm}
\usepackage{mathtools}  %
\usepackage{adjustbox}
\newtoggle{arxiv}
\toggletrue{arxiv}

  \usepackage[parfill]{parskip}
  \usepackage[citestyle=authoryear-comp, uniquename=false, uniquelist=false,sorting=ynt, maxbibnames=99, maxcitenames=1, backend=biber]{biblatex}
  \newcommand{\citep}{\parencite}
  
  \newcommand{\citet}{\textcite}
  \newlength{\defbaselineskip}
  \RequirePackage[T1]{fontenc}
  \RequirePackage[tt=false, type1=true]{libertine}
  \RequirePackage[varqu]{zi4}
  \RequirePackage[libertine]{newtxmath}

\newcommand{\SQuAD}{\textsc{SQuAD}\xspace}
\usepackage{xspace}

\usepackage{bm}

\usepackage[inline]{enumitem}
\usepackage{booktabs}
\usepackage{subcaption}
\usepackage{multirow}
\usepackage{wrapfig}
\usepackage{algorithm,algorithmicx,algpseudocode}
\usepackage{nicematrix}

\usepackage[frozencache,cachedir=.]{minted} %

\usepackage{import}
\usepackage{lettrine}

\newtheorem{dfn}{Definition}

\usepackage{pifont}%
\usepackage[T1]{fontenc}

\usepackage{thm-restate}

\usepackage{url}            
\usepackage{booktabs}       
\usepackage{amsfonts}       
\usepackage{nicefrac}       
\usepackage{microtype}      

\usepackage{wrapfig}

\usepackage{caption}
\usepackage{capt-of}
\usepackage{lipsum}

\usepackage[colorlinks]{hyperref}
\usepackage[capitalise,noabbrev]{cleveref}

\usepackage{threeparttable}

\definecolor{c7}{HTML}{6C7869}
\definecolor{c6}{HTML}{FBE7B4}
\definecolor{c5}{HTML}{9B5353}
\definecolor{c1}{HTML}{586770}
\definecolor{c4}{HTML}{2a4a67}
\definecolor{c3}{HTML}{6d2a58}
\definecolor{c2}{HTML}{34142a}
\hypersetup{colorlinks=true, citecolor=c3, linkcolor=c2, urlcolor=c2}

\definecolor{mydarkred}{HTML}{9B5353}
\definecolor{mydarkpurple}{HTML}{65657E}
\definecolor{mydarkgreen}{HTML}{6C7869}
\definecolor{darkred}{HTML}{a70c0c}
\definecolor{darkblue}{HTML}{4D5B9C}

\definecolor{myblue}{HTML}{FDF5E0} 
\definecolor{mygray}{HTML}{EDECE6} 
\definecolor{mygreen}{HTML}{E6F3FC}

\definecolor{dark2orange}{rgb}{0.9, 0.4, 0.}
\definecolor{dark2purple}{rgb}{0.4, 0.4, 0.8}

\hypersetup{
    colorlinks=true,
    linkcolor=black,
    urlcolor=c1,
    citecolor=c1,
}

\usepackage{multirow}

\newcommand{\R}[0]{\mathbb{R}}

\newcommand{\E}{\mathbb{E}}

\def\D{\mathcal{D}}

\newcommand{\head}[1]{\vspace{1.7mm}\noindent{{\textcolor{c4}{\bf #1.}}}}

\newcommand{\headdot}[1]{\vspace{1.7mm}\noindent{{\textcolor{c4}{\bf #1}}}}

\newcommand{\model}[0]{\textsc{Hope}}

\usepackage{mathtools}

\usepackage{tikz}

\usepackage{authblk}
\usepackage{epigraph}
\usepackage{soul}

\usepackage{dsfont}

\usepackage{tcolorbox}
\tcbuselibrary{breakable}
\usepackage{xcolor}
\usepackage{lettrine,Zallman}
\newtcolorbox{c4box}{boxrule=0.75pt, left=3pt,right=3pt, colback=c4!3!white,colframe=c4!70!white}
\newtcolorbox{c3box}{boxrule=1pt, colback=c3!5!white,colframe=c3!50!white}
\newtcolorbox{c2box}{boxrule=1pt, colback=c2!5!white,colframe=c2!50!white}
\newtcolorbox{c1box}{boxrule=1pt, colback=c1!5!white,colframe=c1!50!white}

\usepackage[framemethod=tikz]{mdframed}
\mdfdefinestyle{mystyle}{%
  rightline=true,
  innerleftmargin=10,
  innerrightmargin=10,
  outerlinewidth=3pt,
  topline=false,
  rightline=true,
  bottomline=false,
  skipabove=\topsep,
  skipbelow=\topsep
}

\newtcolorbox{myboxi}[1][]{
  breakable,
  title=#1,
  colback=c4!5,
  colbacktitle=c4!5,
  coltitle=black,
  fonttitle=\bfseries,
  bottomrule=0pt,
  toprule=0pt,
  leftrule=2pt,
  rightrule=2pt,
  titlerule=0pt,
  arc=0pt,
  outer arc=0pt,
  colframe=c4,
}

\newtcolorbox{myboxismall}[1][]{
  breakable,
  title=#1,
  colback=c5!5,
  colbacktitle=c5!5,
  coltitle=black,
  fonttitle=\bfseries,
  bottomrule=0pt,
  toprule=0pt,
  leftrule=2pt,
  rightrule=2pt,
  titlerule=0pt,
  arc=0pt,
  outer arc=0pt,
  colframe=c5,
}

\newtcolorbox{myboxismallyellow}[1][]{
  breakable,
  title=#1,
  colback=c6!5,
  colbacktitle=c6!5,
  coltitle=black,
  fonttitle=\bfseries,
  bottomrule=0pt,
  toprule=0pt,
  leftrule=2pt,
  rightrule=2pt,
  titlerule=0pt,
  arc=0pt,
  outer arc=0pt,
  colframe=c6,
}

\newtcolorbox{myboxismallgreen}[1][]{
  breakable,
  title=#1,
  colback=c7!5,
  colbacktitle=c7!5,
  coltitle=black,
  fonttitle=\bfseries,
  bottomrule=0pt,
  toprule=0pt,
  leftrule=2pt,
  rightrule=2pt,
  titlerule=0pt,
  arc=0pt,
  outer arc=0pt,
  colframe=c7,
}

\newtcolorbox{myboxnote}[1][]{
  breakable,
  title=#1,
  colback=orange!0,
  colbacktitle=orange!0,
  coltitle=black,
  fonttitle=\bfseries,
  bottomrule=0pt,
  toprule=0pt,
  leftrule=2pt,
  rightrule=2pt,
  titlerule=0pt,
  arc=0pt,
  outer arc=0pt,
  colframe=orange,
}

\newcommand\blfootnote[1]{%
  \begingroup
    \addtocounter{footnote}{3}%
  \renewcommand\thefootnote{}\footnote{#1}%
  \addtocounter{footnote}{-4}%
  \endgroup
}
\newcommand\blfootnotee[1]{%
   \addtocounter{footnote}{0}%
  \begingroup
  \renewcommand\thefootnote{}\footnote{#1}%
  \endgroup
}

\usepackage{authblk}
\usepackage{epigraph}

\usepackage{lettrine,Zallman}

\title{\vspace{-5ex} Language Models Need Sleep: \\Learning to Self-Modify and Consolidate Memories}

\author{Ali Behrouz $^{\dagger}$ \!\!\protect \blfootnote{Correspondence
to: \texttt{\{alibehrouz,~adeljavanmard,~mirrokni\}@google.com} \: and \: \texttt{sh2574@cornell.edu}.} \protect \blfootnotee{A version of this work has been publicly available from September 2025 on OpenReview.}}
\author{Farnoosh Hashemi $^\ddagger$}
\author{Adel Javanmard $^{\dagger}$}
\author{Vahab Mirrokni $^\dagger$}

 \vspace{4ex}

\affil[]{\protect \Large $^{^{^\dagger}}$\includegraphics[width=40mm]{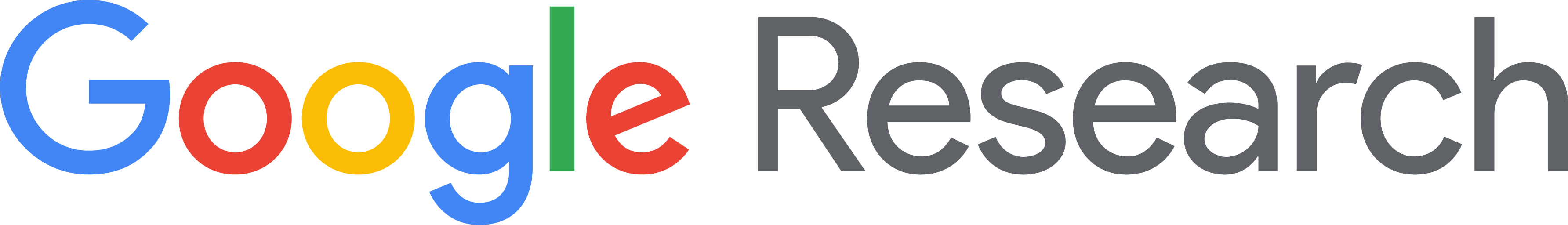} \qquad \qquad \Large $^{^{^\ddagger}}$\protect \includegraphics[width=40mm]{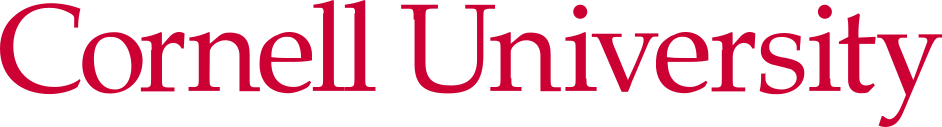}} 
\date{}

\begin{document}

\maketitle

\begin{abstract}
The past few decades have witnessed significant advances in the design of machine learning algorithms–from early studies on task-specific shallow models to more general deep Large Language Models (LLMs). Despite showing promising results in tasks that require instant prediction or in-context learning, existing models lack the ability to continually learn and effectively transfer their temporal in-context knowledge to their long-term parameters. Inspired by human learning process, we introduce a ``\emph{Sleep}'' paradigm that allows the models to continually learn, distill their short-term fragile memories into stable long-term knowledge with replay, and recursively improve themselves with ``\emph{Dreaming}'' process. In more detail, sleep consists of two stages: (1) \textbf{Memory Consolidation}: an \emph{upward} distillation process, called \emph{Knowledge Seeding}, where the memories of a \emph{smaller}-self are distilled into a \emph{larger} network to provide more capacity while preserving the knowledge. As a proof of concept, we present a new \emph{Generalized Distillation} process for {Knowledge Seeding} (i.e., the combination of on-policy distillation with Reinforcement Learning (RL)-based imitation learning); (2)~\textbf{Dreaming}: a self-improvement phase, where the model uses RL to generate a curriculum of synthetic data to rehearse new knowledge and refine existing capabilities without human supervision.  Our experiments on long-horizon, continual learning, knowledge incorporation, and few-shot generalization tasks support the importance of the sleep stage. 
\end{abstract}

\section{Introduction}\label{sec:intro}
The development of Large Language Models (LLMs) marks a pivotal milestone in machine learning research: a paradigm shift from task-specific models to more general-purpose systems with various emergent capabilities~\citep{brown2020language, schaeffer2023emergent}. Despite LLMs' remarkable capabilities in diverse sets of tasks~\citep{wang2023visionllm, nijkamp2023codegen, comanici2025gemini}, they are largely static after their initial deployment, meaning that they successfully perform tasks learned during pre- or post-training, but are unable to \emph{continually acquire} new capabilities beyond their immediate context. This inherent static nature creates a crucial vulnerability: The model's knowledge and skills become progressively stale, operating with a fixed "knowledge cutoff" date beyond which it is unaware of new facts, events, and evolving information~\citep{cheng2024dated}. 

Efforts to overcome this limitation have primarily focused on: (1) re-pretraining on an expanded dataset, which despite its effectiveness, is computationally expensive and impractical for frequent updates~\citep{ibrahim2024simple}; (2) using expensive continual parameter updates or other lightweight alternatives, such as fine-tuning or low-rank adaption~\citep{hu2022lora, akyureksurprising}, which with iterative updates often results in Catastrophic Forgetting (CF)~\citep{kemker2018measuring, shi2024continual}–a well-known phenomenon where the model's proficiency on original tasks degrades catastrophically as it learns new ones. This dilemma—between knowledge obsolescence on one hand and catastrophic forgetting as well as the prohibitive cost or destructive nature of updates on the other—underscores a critical, unresolved challenge: enabling LLMs to learn incrementally and efficiently throughout their lifecycle.

\begin{figure*}
    \centering
    \includegraphics[width=\linewidth]{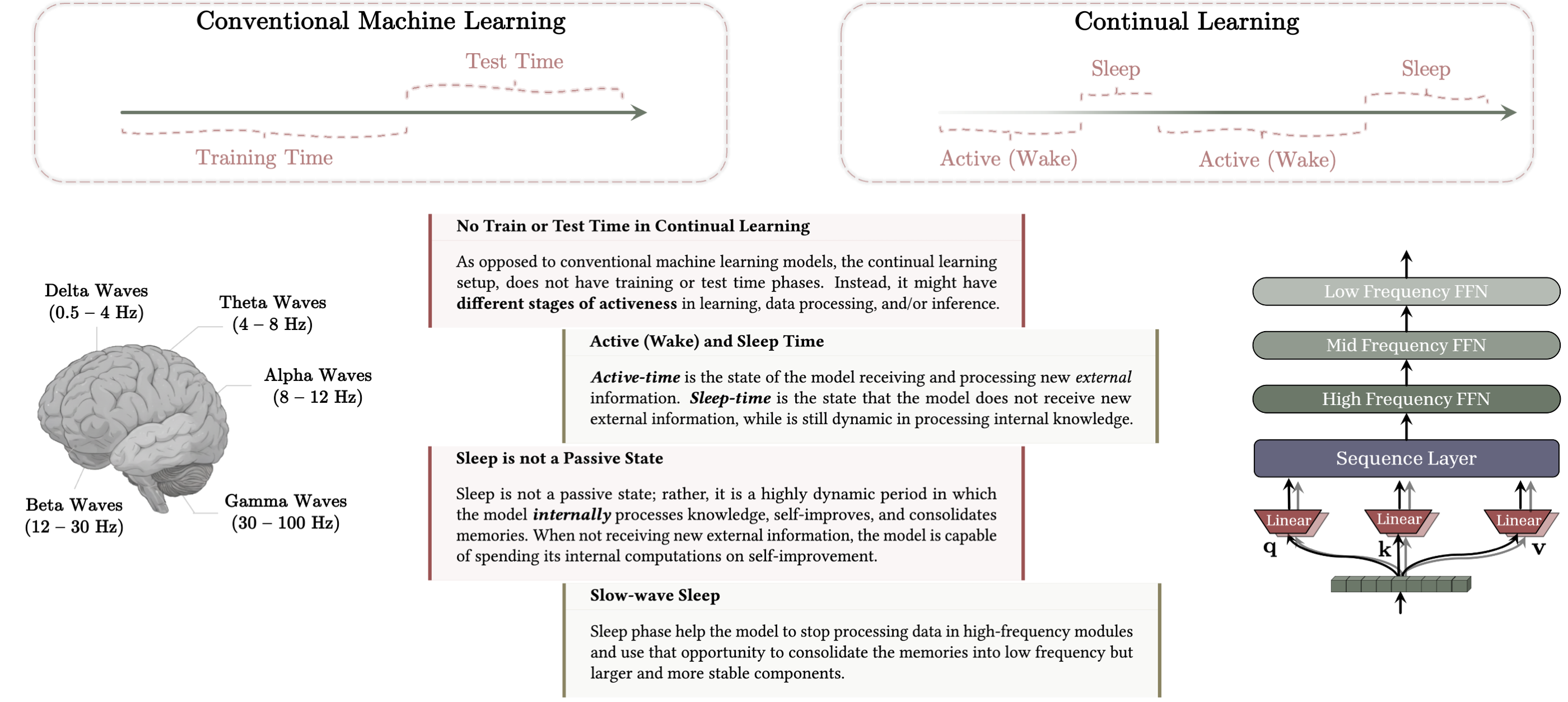}
    \caption{(\textbf{Conventional Machine Learning vs. Continual Learning}) While in conventional machine learning often the lifespan of the model is divided to \emph{test and training time}, continual learning setup does not have these phases. We suggest that a continual learner need to have different stages of activeness in learning, which we refer to as: (i) \emph{Active or Wake Time}, and (ii) \emph{Sleep Time}. Sleep time is not a passive state, rather it internally process the data to consolidate the memories from fast unstable modules to more stable low frequency (slow) components.}
    \label{fig:sleep-states}
\end{figure*}

In recent years, In-Context Learning (ICL)~\citep{brown2020language} has gained attention as a highly efficient and successful form of continual learning~\citep{akyurek2022learning, dong2024survey, akyurek2024context, li2025longcontext}. Initially, ICL was known as an emergent ability of LLMs that is trained on large scale data, enabling them to adapt fast to the context and so perform zero- or few-shot tasks~\citep{brown2020language}. Later, more studies revealed and formalized the role of ICL as a meta-learning process in which the model performs internal computations along the sequence to incorporate context knowledge to its output by keeping or compress it into a short-term memory~\citep{behrouz2025nested, dherin2025learning}. Despite the effectiveness/efficiency of ICL as a form of continual learning, it is limited to the context-window of sequence models, meaning that any new acquired knowledge will be removed from the model at the end of the session/context. This perspective raises a critical question: \emph{How the model can effectively transfer the fragile short-term memories into more stable long-term knowledge}?

As an analogy, we use the example of anterograde amnesia for LLMs from \citet{behrouz2025nested}: Consider an impairment in the process of transferring the information from short-term to longer-term memories in humans, example of which is anterograde amnesia–a neurological condition where a person cannot form new memories after the onset of the disorder, while existing memories remain intact~\citep{scoville1957loss}. Such conditions can limit the person's knowledge to immediate present that fits in the short-term memory and long past, before the onset of the disorder, resulting in continuously experiencing the immediate present as if it were always new. One might notice a similar pattern in the memory processing of current Transformer-based LLMs. The knowledge of LLMs are limited in either: (1) the immediate context that fits into their context window (a.k.a. in-context learning), or (2) MLP and projection layers, storing long-past, before the onset of ``end of pre-training.'' This similarity in pattern motivates us to ask, \emph{What is the critical component in human learning process that consolidates memories?}

The human brain is highly efficient, lossy but effective when it comes to consolidating memory, which is often attributed to neuroplasticity—the brain's ability to change itself in response to new experiences, memories, and learning~\citep{pascual2005plastic, johnston2009plasticity}. Recent studies support that long-term memory formation involves at least two distinct but complementary consolidation processes~\citep{Stepwise-consolidation2021, frey1997synaptic, yang2024selection}: (1) A rapid ``online'' consolidation phase occurs immediately or soon after learning, even during wakefulness. This is when new fragile memory traces are stabilized and begin transferring from short-term to more longer-term storage; (2) An ``offline'' consolidation (also known as systems consolidation) process repeats the replay of the recently encoded patterns during sleep and reorganizes the memory and supports transfer to cortical sites~\citep{ji2007coordinated, peyrache2009replay, foster2006reverse}.

\begin{myboxismall}[Online Consolidation in Humans (in Active Stage)]
\textit{Online consolidation in humans is the active, \underline{wake-state} strengthening and transformation of a newly learned memory by reactivating the memory during recall, making it more stable and more semantic-like over time.}
\end{myboxismall}

\begin{myboxismallyellow}[Offline Consolidation in Humans (in Sleep and Quiet Rest Stages)]
\textit{Offline consolidation in humans occurs \underline{after learning} during rest or sleep, when a newly encoded memory is stabilized and gradually transformed into a more distributed neocortical representation by a complex lossy distillation process.}
\end{myboxismallyellow}

Returning to the analogy of anterograde amnesia, the evidence indicates that the condition can affect both stages. \citet{behrouz2025nested}, recently, presented the Nested Learning (NL) paradigm and aimed at the first form of memory consolidation (i.e., online consolidation). In particular, NL-based Hope architecture (see \autoref{fig:sleep-states}) with the reactivation of the memory in a continuum memory system, where each component is updated with a different frequency but in an end-to-end manner, transfers the knowledge from fast unstable components to more stable low frequency modules in an online manner. Although this end-to-end process can directly transfer the knowledge to more stable components and so \emph{postpone} forgetting, it still uses the same amount of model's capacity. This mainly happens due the fact that the model does keep the knowledge at the same level of abstraction without any additional lossy compression. Furthermore, this form of consolidation is not enough for robust continual learning: \textbf{(i)} Is selective and retrieval-dependent: it depends on active retrieval and so strengthens the parts of a memory that are repeatedly recalled; \textbf{(ii)} Depends on the context: the update caused by the online consolidation is based only on the context of the model, and so misses the higher-level understanding of new knowledge with existing one. In this paper, we focus on the second type of consolidation: i.e., offline consolidation via sleep, which is inspired by the sleep stage in humans:

\subsection*{The Role of Sleep in Human Learning Process}
Sleep is not a passive state but a dynamic and highly structured period of brain activity essential for cognitive function~\citep{rasch2013sleep, goldstein2014role}. During sleep, the brain orchestrates complex processes fundamental to learning, neural plasticity, self-improvement, and memory consolidation~\citep{wamsley2011memory, rasch2013sleep, goldstein2014role}. In humans, these processes are primarily governed by two critical and alternating stages of sleep: Rapid Eye Movement (REM) and Non-REM (NREM) sleep.

\headdot{Non-Rapid Eye Movement Sleep (Slow-Wave Sleep):} This stage, particularly its deepest phase known as slow-wave sleep, is characterized by synchronized, high-amplitude, low-frequency neural activity. Slow-wave sleep is associated with two primary functions crucial for learning: The first is synaptic homeostasis, a process that globally downscales synaptic strengths to counteract the net increase in connectivity from waking experiences, thereby maintaining metabolic balance and preventing neural saturation~\citep{tononi2006sleep}.

The second core function is memory consolidation, the transformation of fragile, recent experiences into stable, long-term knowledge~\citep{squire1995retrograde}. This process is orchestrated through a sophisticated dialogue between the hippocampus and the neocortex~\citep{squire2015memory}. The hippocampus serves as a high-fidelity temporary storage system, capable of rapidly encoding specific daily experiences. In contrast, the neocortex is a vast, long-term repository better suited for the gradual learning of generalized rules and semantic knowledge from these experiences~\citep{squire1995retrograde, squire2015memory}. During slow-wave sleep, the brain initiates a nightly dialogue between these structures that facilitates an intricate transfer of information. Notably, this transfer does not simply replay raw data; instead, it re-architects the knowledge acquired during waking hours, extracting abstractions and integrating them into a cohesive semantic network.

\headdot{Rapid Eye Movement Sleep:} Characterized by high-frequency, low-amplitude brain waves that resemble an awake state, REM sleep is most commonly associated with dreaming. Functionally, this stage is linked to the selective strengthening of newly formed synapses and the integration of new information with pre-existing emotional and semantic networks. Furthermore, it is hypothesized to play a role in simulating future scenarios to improve adaptive behavior.

In summary, the cyclical alternation between NREM and REM stages throughout the night is crucial. NREM sleep appears to consolidate and prune the day's experiences to build a more efficient knowledge base. Subsequently, REM sleep seems to operate on this refined base, exploring novel connections and strengthening salient neural pathways.

\begin{myboxismall}[No Training or Test Time in Continual Learning]
    \textit{For a continual learner, there is no boarder and clear distinction between training and test time. The model only experiences two different states: when it receives information as input, or when it is an isolated learning system.}
\end{myboxismall}

\begin{myboxismallyellow}[Active (Wake) and Sleep Phases in Continual Learning]
    \textit{In the active or wake time, continual learner receives new input data and process it as needed. In the sleep time, however, the model receives minimal (or none) input data and focus on processing existing knowledge to consolidate memories and self-improve. }
\end{myboxismallyellow}

\subsection*{Contributions} 
Inspired by the memory processing in humans, we argue that for a continual learner, there is no train or test time. Instead, the model needs to periodically have two phases of being "active" or "Sleep"; In the active state, the model receives new data and processes it, while in the sleep phase the focus is on the internal knowledge and the consolidation of recent memories.  To this end, we introduce an instance of ``sleep'' paradigm for LLMs that consists of two integrated phases: 
\begin{enumerate}[label=\roman*]
    \item \textbf{Memory Consolidation}: To mitigate catastrophic forgetting (CF) and to capture higher-levels of abstraction, in the memory consolidation phase, the model uses a periodic process in which it activates/unlocks new parameters and distills the knowledge from higher-frequency (i.e., faster updating) layers/modules to the newly unlocked parameters in more stable (lower-frequency) layers/modules. This process allows enough plasticity for new parameters while ensuring the stability of old parameters, preserving the old knowledge.
    \item \textbf{Self-Improvement via Dreaming}:  While the previous first stage of sleep ensures transferring the knowledge abstraction to longer-term memories, this stage is responsible for the process of recursive self-improvement. In particular, given the current state of the model, it generates a set of dreams (i.e., synthetically self-generated data) to improve its own performance with particular focus on the acquiring more proficiency on the recently added knowledge. 
\end{enumerate}
From the technical point of view, our contributions to each of the above phases are:
\begin{enumerate}
    \item \textbf{Periodic Parameter (De)Activation}: Building on the Nested Learning (NL) paradigm~\citep{behrouz2025nested} that allows each component to have its own frequency of update, we suggest a periodic and gradual parameter (de)activation process, where given a block and for each sleep step, we deactivate a set of parameters in a faster block (i.e., higher frequency block) and replace them with a set of newly activated parameters in the current block. This allows maintaining plasticity while avoiding knowledge interference with previous parameters. 
    \item \textbf{Knowledge Seeding (Upward Distillation)}: We present a new form of knowledge transfer, called knowledge seeding, where one or some \textbf{\emph{smaller}} models distill their knowledge to a \textbf{\emph{larger}} model. This design allows the larger model to preserve existing knowledge in smaller models, while taking advantage of its larger capacity. Based on the formulation of Knowledge Seeding (KS), we present self-Knowledge Seeding (SKS), where a smaller version of a model (e.g., some parameters are not active), distill the knowledge to a larger version of the model (e.g., where parameters are active). We then use SKS as a solution for memory consolidation, in which the model distills its knowledge from high-frequency layer/blocks to low-frequency blocks. 
    \item \textbf{Generalized Knowledge Distillation (GKD) with Imitation Learning}: The formulation of SKS is general, and any form of objective and knowledge transfer method can be used. Here, we present a new objective that combines on-policy distillation with an imitation learning process. In particular, in our imitation learning process, we suggest that the teacher (i.e., a smaller version of model with (compressed) privilege information), generates synthetic data and then masks the sequence; Then, the student aims to predict the continuation of the sequence and gets its reward based on the distance of teacher generated data and its own prediction.  
    \item \textbf{Random Expert Selection for Dreaming}: The dreaming phase is responsible for understanding the relationship between different types of knowledge, and so we present a random low-rank expert selection in Mixture of Experts (MoE)~\citep{wu2024mixture} that aims to mix the knowledge by combining the knowledge of irrelevant experts for output computation. 
    \item \textbf{Experimental Evaluation}: We evaluate the effectiveness of sleep paradigm on a set of challenging downstream tasks: (1)  Factual Knowledge Incorporation; (2) Few-shot Learning; (3) Long-context Understanding; and (4) Continual Learning. The results support the effectiveness of Sleep paradigm as well as the importance of growing parameters with iterative knowledge distillation for continual learning.
\end{enumerate}

\section{Preliminaries and Problem Formulation}  \label{sec:prelim}

\subsection{Notation} \label{sec:notation}
We use bold lowercase (resp. uppercase) letters for vectors (resp. matrices) and use subscript $t$ to refer to the state of the entities correspond to time $t$. Superscripts for parameters of a module (resp. hyperparameters) are used to determine the update frequency of the module (resp. distinguish different instances). Through the paper, we let $x \in \R^{L \times d_{\text{in}}}$ be the input, $\mathbf{K}$ be the keys, $\mathbf{V}$ be the values, $\mathbf{Q}$ be the query matrices in the sequence model, and $L$ denote the sequence length. When it is needed, we parameterize the language model $\texttt{LM}_{\theta}$ with $\theta = \{W^{(f_1)}_{1}, \dots, W_{k_1}^{(f_{1})}\} \cup \{W^{(f_2)}_{1}, \dots, W_{k_2}^{(f_{2})}\} \cup \dots \{W^{(f_c)}_{1}, \dots, W_{k_c}^{(f_{c})}\}$, where parameter sets are sorted based on their weight update frequencies $f_1 \geq \dots, \geq f_c$ (see Definition~\ref{dfn:freq}).

\subsection{Continuum Memory System}  \label{sec:CMS}
Transformer architectures consist of two critical components: (1) Attention module that acts as an associative memory and conditions the output on the past tokens in the context, which also results in in-context learning ability; and (2) MLP or feedforward layers, which are fixed after the training phase and encodes the knowledge acquired over the pre-training. As discussed by \citet{behrouz2025nested}, one can interpret such architectures as two-level memory systems, in which the attention's update span is the context length–meaning that at the end of the context, its corresponding parameters are updated and the acquired knowledge is forgotten–and MLP's update span is non-existence–indicating no update after pre-training. From this perspective, these two components are two extreme sides of the frequency spectrum, where the attention (resp. MLP) has infinite (resp. zero) update frequency.

Building on this intuition, \citet{behrouz2025nested} presented Continuum Memory System (CMS), where the architecture is a sequence model such as attention, followed by a chain of MLP layers, each of which updated with its own frequency. More specifically, the time for one step of update in the slowest module is considered as the unit of time, and so the update rate of other components are defined as: 

\begin{dfn}[Update Frequency]\label{dfn:freq}
    For any weight component of $W$, we define its frequency, denoted as $f_W$, as its number of updates per unit of time.   
\end{dfn}
To better understand this concept, we use a simple example of Fast-weight Programs~\citep{schmidhuber1992learning}, where the input is a sequence of length $L$. In this case, for each step of slow-weight (the unit of time), the fast-weight is updated $L$ times, resulting in an update frequency of $L$. 

Following this definition of frequency, which at high-level indicates how often the parameters of a module are updated over time, CMS is formalized as a chain of MLP blocks $\texttt{MLP}^{(f_1)}(\cdot), \dots, \texttt{MLP}^{(f_k)}(\cdot)$, each of which is associated with a chunk size of $C^{(\ell)} := \frac{\max_{\ell'} C^{(\ell')}}{f_{\ell}}$ such that given input $x = \{x_1, \dots, x_T\}$, {with each $x_t$ as one of the input data,} the output of the chain is calculated as (we disregard normalizations for the sake of clarity):
\begin{align}
    y_t = \texttt{MLP}^{(f_k)}(\texttt{MLP}^{(f_{k-1})}(\cdots \texttt{MLP}^{(f_1)}(x_t))), 
\end{align}
where the parameters of $\ell$-th MLP block, i.e., $\boldsymbol{\theta}^{(f_{\ell})}$, are updated every $C^{(\ell)}$ steps: i.e., $\boldsymbol{\theta}^{(f_{\ell})}_{i+1} = \boldsymbol{\theta}^{(f_{\ell})}_i - \boldsymbol{e}_{i, \ell}$ where:
\begin{align}\label{eq:CMS}
    \boldsymbol{e}_{i, \ell} = \begin{cases} \sum_{t = i - C^{(\ell)}}^{i}\eta^{(\ell)}_t f(\boldsymbol{\theta}^{(f_{\ell})}_{t}; x_{t}) & \text{if \:} i \equiv 0 \:\: (\texttt{mod} \: C^{(\ell)}), \\ 0 & \text{otherwise}.
    \end{cases}
\end{align}
Here $\eta^{(\ell)}_t$ are learning rates corresponds to $\boldsymbol{\theta}^{(f_{\ell})}$, and $f(\cdot)$ is the error component of an arbitrary optimizer (e.g., $\nabla \mathcal{L}(\boldsymbol{\theta}^{(f_{\ell})}_{t}; x_t)$ in gradient descent) in an end-to-end optimization of the neural network. The formulation of CMS is very broad and, depending on the task, the objective $\mathcal{L}(\cdot)$ can be changed (the same as the MLP blocks in Transformers). Due to this generality of formulation and being the superset of Transformers design (i.e., 1 MLP blocks will be equivalent to Transformers), throughout the paper, we use CMS as the default building blocks of the architectures. {Also, for the sake of clarity and without loss of generality, we assume that $C^{(\ell)}$ is divisible by $C^{(\ell - 1)}$}. It is notable that \autoref{eq:CMS} provides an important interpretation: parameters $\boldsymbol{\theta}^{(f_{\ell})}_{t}$ are responsible for compressing their own context into the their parameters and so they are a representative of abstract knowledge of their context (We refer to the Nested Learning paper~\citep{behrouz2025nested} for more details).       

In summary, in this perspective, the sequence model (e.g., attention~\citep{transformers} or other memory modules or RNNs~\citep{katharopoulos2020transformers, behrouz2024titans}) acts as the short-term memory of the model since their high-frequency update can push the old knowledge to be forgotten, making space for new memories. On the other hand, CMS blocks act as a spectrum of memory modules, in which earlier blocks (higher-frequency) are shorter-term memories, while later blocks (and ultimately the last one with close to zero frequency) are longer-term memories. While actively updating this memory system can enhance the resistance to CF, the CF can happen when the update period of all models matched at some point~\citep{behrouz2025nested}. Therefore, it is crucial that before each update of a memory block, a mechanism consolidates the abstracted knowledge of that block to more stable parameters.  

\subsection{Sleep Terminology in the Literature} 
The human sleep process has been an inspiration for the design of many studies in the literature~\citep{mcclelland1995complementary, kumaran2016learning, hassabis2017neuroscience, ha2018world}. In particular, ``dreaming'' has motivated several past studies in the literature to design methods that replay recent experiences/input data~\citep{lin1992self, mnih2015human, ha2018recurrent, ha2018world}. Several studies designed an offline process that can make the model more robust for long-horizon tasks~\citep{hafner2020dream, tadros2022sleep, gonzalez2020sleep}. \citet{lin2025sleep} suggest an offline self-study process as a form of ``sleep-time compute'' that summarizes the past context for the next user session, mainly is designed as a proposal for spending compute on the context rather than training the model itself.

To the best of our knowledge, the existing literature (including sleep-inspired studies) remains firmly anchored in the conventional distinction of training and testing phases. Even in the continual or online learning setups, models alternate between updating parameters (training) and evaluating performance (testing). We argue that for lifelong adaptation, the static train/test paradigm needs to be replaced with a continuous periodic "wake" and "sleep" lifecycle. During the "wake" phase, the model is actively interacting with varying input data and rapidly acquiring temporary information while during the "sleep" phase, the model consolidates its memory and internally processes existing knowledge. 
\section{The Sleep Paradigm}  \label{sec:sleep-paradigm}

\subsection{Learning Phases in Continual Learning}\label{sec:continual-learning-setup}
As discussed earlier, a continual learner is always learning from the data/experience. Therefore, the conventional split of the machine learning models' lifecycle (i.e., test and train time) is not directly applicable in the continual learning setup. We suggest the use of ``active or wake'' time and ``sleep'' time as the two critical phases of a lifecycle of a continual learner. In particular, in the active or wake time, continual learner receives new input data and process it as needed. In the sleep time, however, the model receives minimal (or none) input data and focus on processing existing knowledge to consolidate memories and self-improve.

However, the process of memory consolidation is not limited to the sleep phase. In fact, following the discussion in Nested Learning (NL)~\citep{behrouz2025nested}, there are two forms of memory consolidations: (1) Online consolidation; and (2) Offline consolidation:

\begin{myboxismall}[Online Consolidation in Neural Networks in Active (Wake) Stage]
Online consolidation happens through the end-to-end learning process of neural learning modules (e.g., Hope architecture in \autoref{fig:sleep-states}) in the wake or active phase, where earlier layers (faster blocks with higher frequency) transfer their knowledge to later layers (i.e., slower blocks with lower frequency). 
\end{myboxismall}

\begin{myboxismallyellow}[Offline Consolidation in Neural Networks in Sleep Stage]
In offline consolidation phase, the model enters the sleep phase, where it stops processing new data, but uses self-generated data to consolidate the recent memories. 
\end{myboxismallyellow}

In the next section, we present Sleep paradigm, in which contrary to the model's waking time (or active time), the model does not receive any external input data and concentrates its internal computations on self-improvement, consolidating the past memories, and abstracting knowledge. In particular, we divide the sleep process into two key stages: (1) Memory consolidation; and (2) Dreaming for self-improvement.

\begin{figure*}
    \centering
    \includegraphics[width=\linewidth]{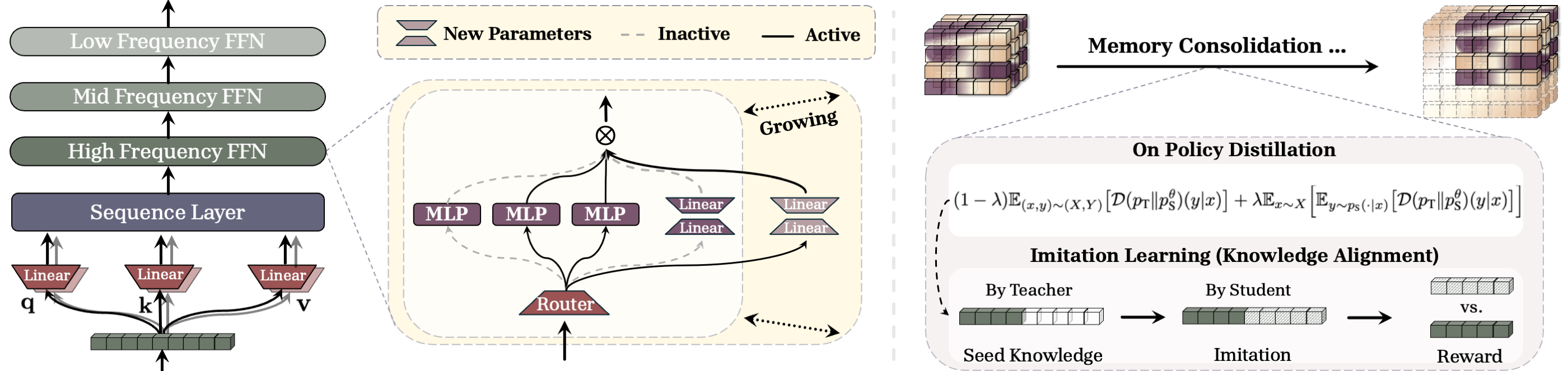}
    \caption{An overview of Memory Consolidation. The model increases its own number of parameters to enhance its capacity (\autoref{sec:param-exp}), and then using the knowledge seeding, it transfers the knowledge abstractions from the higher to a lower-frequency memory (\autoref{sec:ks}).}
    \label{fig:consolidation}
\end{figure*}

\subsection{Memory Consolidation: Parameter Expansion}\label{sec:param-exp}
As discussed earlier, in memory consolidation, we aim to transfer the short-term fragile memories into more vast and stable parameters. One of the important messages in CMS formulation is: the fragility and/or stability of memories are relative. That is, for each memory block, other higher frequency memories are shorter-term and more fragile. Therefore, memory consolidation is not a simple two-step process, but an iterative operation that repeatedly transfers the knowledge stored in higher frequency memories into more stable lower-frequency parameters.

To avoid losing the knowledge of a faster updating block; an example of which is in-context learning~\citep{brown2020language}, we need to perform memory consolidation step before updating its parameters. Therefore, given the list of chunk lengths $\{C^{(1)}, \dots, C^{(k)}\}$, the sleep process and so memory consolidation happens only at $\{C^{(1)} \times b, \dots, C^{(k)} \times b\}$ steps for all $b \in \mathbb{N}$. Based on the update frequency of MLP blocks, we might need to consolidate the memory of a memory module into its next memory block multiple times. For example, consider a memory with update frequency of 1K followed by a memory with update frequency of 10K: in this case, the faster memory is updated 10 times before the update of slower memory block, which means 10 memory consolidation steps of faster memory to slower memory before slower memory's own update. This multiple consolidation steps into the slower memory can be a critical bottleneck to unlock continual learning capabilities of LLMs, due to the Catastrophic Forgetting (CF). This phenomenon is an inherent cause of model's limited capacity (e.g., number of parameters), where parameters need to be overridden to incorporate the new knowledge. The foundation of human's brain solution to this challenge is neuroplasticity, the brain's inherent ability to modify its own function and shape new connections in response to experiences. Inspired by this, we present an efficient gradual parameter expansion in memory blocks that allows the model to shape new connections and so increases its own capacity. 

Without loss of generality, we assume that the MLP blocks $\{\texttt{MLP}^{(f_{\ell})}(\cdot)\}_{\ell = 1}^{k}$ are sparse mixture of experts (MoEs) with a router $\mathcal{R}^{(f_{\ell})}$: i.e., each $\texttt{MLP}^{(f_{\ell})}(\cdot)$ includes a set of experts $\{W^{(f_{\ell}), 1}, \cdots, W^{(f_{\ell}), \mathbf{s}_{\ell}} \}$, where $\mathbf{s}_{\ell} \geq 1$ is the current number of experts in the $\ell$-th block of the chain. Let $(\ell^{*} - 1)$ be the index of the memory (or MLP) that we aim to consolidate its knowledge to its immediate next more stable memory module with index $\ell^{*}$. To avoid the interference of transferred and previously stored knowledge in $\texttt{MLP}^{(f_{\ell^*})}(\cdot)$, we add a new low-rank expert to its set of parameters. That is, we add a low-rank MLP parametrized by $\{\textbf{A}^{(f_{\ell^*}), \mathbf{s}_{\ell^*} + 1}, \textbf{B}^{(f_{\ell^*}), \mathbf{s}_{\ell^*} + 1} \}$, where $\textbf{A}^{(f_{\ell^*})} \in \mathbb{R}^{d \times d_{\text{low}}}$ and $\textbf{B}^{(f_{\ell^*})} \in \mathbb{R}^{d_{\text{low}} \times d}$ ($d_{\text{low}} \ll d$), to the set of experts. These new parameters will be allocated for storing the new transferred knowledge from $\texttt{MLP}^{(f_{\ell^* - 1})}(\cdot)$. Given this process, after each sleep time, the parameters of a subset of layers are growing.

\subsection{Memory Consolidation with Knowledge Seeding}  \label{sec:ks}

{In this step, we aim to transfer the knowledge of $\texttt{MLP}^{(f_{\ell^* - 1})}(\cdot)$ with parameters $\boldsymbol{\theta}^{(f_{\ell^* - 1})}$ into the expanded set of parameters in $\texttt{MLP}^{(f_{\ell^*})}(\cdot)$. Since sleep is triggered when the number of past steps is divisible by $C^{\ell^* - 1}$, the sender block $\texttt{MLP}^{(f_{\ell^* - 1})}(\cdot)$ is scheduled for a base-weight update at this time. The consolidation procedure follows a compute–consolidate–update protocol:
\begin{itemize}
\item[$(a)$] \emph{Compute.} The base-weight update for the sender is computed from the accumulated gradients (Equation~\eqref{eq:CMS}), yielding prospective new parameters $\boldsymbol{\theta}^{(f_{\ell^* - 1})}$. Crucially, this update is computed but not yet applied to the running model.

\item[$(b)$] \emph{Consolidate.} We define:
\begin{itemize}
\item {\bf Teacher ($\texttt{LM}_{\boldsymbol{\theta}}$):} the state of the model before the base-weight update or parameter expansion — i.e., $\texttt{MLP}^{(f_{\ell^*})}(\cdot)$ still uses its original parameters $\boldsymbol{\theta}^{(f_{\ell^* - 1})}$ plus its accumulated experts;
\item {\bf Student ($\texttt{LM}_{\boldsymbol{\theta}_{\text{exp}}}$):} the state of the model using the prospective updated parameters  $\boldsymbol{\theta}^{(f_{\ell^* - 1})}_{\text{new}}$ (with experts in $\texttt{MLP}^{(f_{\ell^* - 1})}(\cdot)$ reset), and with a newly added low-rank expert {A, B} in $\texttt{MLP}^{(f_{\ell^*})}(\cdot)$.
\end{itemize}
The low-rank parameters $\{A, B\}$ are then optimized via the Knowledge Seeding objective (defined below) to minimize the distillation loss between teacher and student.

\item[$(c)$] {\emph Update.} After the optimal {A, B} are found: (a) the base-weight update $\boldsymbol{\theta}^{(f_{\ell^* - 1})} \to \boldsymbol{\theta}^{(f_{\ell^* - 1})}_{\text{new}}$ is applied, (b) the low-rank experts previously added to $\texttt{MLP}^{(f_{\ell^* - 1})}(\cdot)$ are reset (synaptic pruning), and (c) the new expert $\{A, B\}$ is activated in $\texttt{MLP}^{(f_{\ell^*})}(\cdot)$.
\end{itemize}
This ordering ensures that the prospective weights $\boldsymbol{\theta}^{(f_{\ell^* - 1})}_{\text{new}}$ are available for defining the student model at step 2, while the teacher still faithfully reflects the pre-update state. We model this as a distillation problem: transferring the knowledge stored in the smaller-state teacher $\texttt{LM}_{\boldsymbol{\theta}}$ to the larger-capacity student $\texttt{LM}_{\boldsymbol{\theta}_{\text{exp}}}$.}

\head{Knowledge Seeding}
We present a new form of knowledge transfer, called knowledge seeding, where one or some \textbf{\emph{smaller}} models distill their knowledge to a \textbf{\emph{larger}} model. This design allows the larger model to preserve existing knowledge in smaller models, while taking advantage of its larger capacity. 

\begin{myboxismall}[Upward Distillation (Knowledge Seeding)]
    Given a set of small models $\mathcal{S}_1(\cdot), \dots, \mathcal{S}_k(\cdot) $ as the teacher, Knowledge Seeding (KS) aims to transfer the knowledge in these models to a larger model $\mathcal{M}(\cdot)$.
\end{myboxismall}

Based on the formulation of Knowledge Seeding (KS), we present self-Knowledge Seeding (SKS), where a smaller version of a model (as discussed above) distills the knowledge to a larger version of the model. This distillation process has two critical challenges: (1) Contrary to conventional cases, student has more capacity and so more expressive power than the teacher. Therefore, training the student on the teacher generated dataset (e.g., \citet{kim2016sequence}) can result in sub-optimal use of parameters in student model; (2) The model is in sleep stage and so the access to the external information/dataset is limited. Therefore, most popular methods like \citet{hinton2015distilling} are not applicable. To overcome these challenges, we build upon Generalized Knowledge Distillation (GKD)~\citep{agarwal2024onpolicy}, which allows a mixture of on-policy student generated data with a teacher-generated data, and present a novel distillation process based on imitation learning. 

As discussed earlier, the memory consolidation step should not simply replay raw data; instead, it needs to explore and extract abstractions of knowledge acquired during active (waking) steps. To this end, knowledge seeding has two main steps: (1) A distillation process, in which student receives token-specific feedback from the teacher’s logits on the self-generated sequences; and (2) An RL-based imitation learning method that forces the student to memorize the sampled outputs of teacher, aligning their sampling process while preserving the distilled knowledge. 

We start with constructing a dataset $\mathcal{D}$ by sampling from the teacher model, i.e., $\texttt{LM}_{\boldsymbol{\theta}}$. Next, similar to GKD~\citep{agarwal2024onpolicy}, we define on policy distillation objective as:
\begin{align}\nonumber
&\mathcal{L}(\boldsymbol{\theta}, \boldsymbol{\theta}_{\text{exp}})
= (1 - \lambda)\,
\E_{(x, y) \sim \mathcal{D}}
\Big[
\mathcal{F}(
\texttt{LM}_{\boldsymbol{\theta}}
\| 
\texttt{LM}_{\boldsymbol{\theta}_{\text{exp}}}
)(y|x)
\Big] + \lambda\,
\E_{x \sim \mathcal{D}}
\Big[
\E_{y \sim \texttt{LM}_{\boldsymbol{\theta}_{\text{exp}}}(\cdot|x)}
\big[
\mathcal{F}(
\texttt{LM}_{\boldsymbol{\theta}}
\|
\texttt{LM}_{\boldsymbol{\theta}_{\text{exp}}}
)(y|x)
\big]
\Big].
\end{align}
where $\mathcal{F}(\texttt{LM}_{\boldsymbol{\theta}}, \texttt{LM}_{\boldsymbol{\theta}_{\text{exp}}})(y|x)$ is a divergence between teacher (i.e., $\texttt{LM}_{\boldsymbol{\theta}}$) and student (i.e., $\texttt{LM}_{\boldsymbol{\theta}_{\text{exp}}}$) output distributions, and $\lambda \in [0, 1]$ controls the fraction of on-policy student-generated outputs. In this optimization process, we do not backpropagate through the sampling distribution of the student, which can help with training stability and also speed. Also, we freeze all the parameters in the student model and only updates the expanded parameters. This ensures that the transferred knowledge does not interfere with the old knowledge, causing catastrophic forgetting.

\head{Learning to Imitate} 
The above distillation process ensures that the student new parameters store the knowledge encoded in the lower-frequency memory. However, we observe that despite having access to the knowledge, the student model has not learned to use it and so weakly mimics the sampling and performance of the teacher. To this end, we further improve the above distillation process by incorporating RL to teach model how to imitate the teacher sampling. Given a set of teacher generated data (dreams), $\mathcal{D}_{T} = \{d^{(1)}, \dots, d^{(n)} \}$, Learning to Imitate (LTI) process first randomly samples a prefix from each $d^{(i)}$ and then asks the student model to complete the continuation. Given the student responses $\hat{d}^{(i)}$ the assigned reward is defined as:
\begin{equation}
\begin{aligned}
\hspace{-2ex}r(\hat{d}^{(i)}; d^{(i)};\texttt{LM}_{\boldsymbol{\theta}_{\text{exp}}})
&= \gamma \times
r_{\text{sem}}(\hat{d}^{(i)}; d^{(i)};\texttt{LM}_{\boldsymbol{\theta}_{\text{exp}}}) + (1 - \gamma) \times
r_{\text{abs}}(\hat{d}^{(i)}; d^{(i)};\texttt{LM}_{\boldsymbol{\theta}_{\text{exp}}}),
\end{aligned}
\end{equation}

where $r_{\text{sem}}(\cdot;\cdot;\cdot)$ (resp. $r_{\text{abs}}(\cdot;\cdot;\cdot)$) assigns a reward based on the semantic similarity (resp. absolute token-level similarity). For semantic similarity, we use a reward model that is frozen and rewards the student with 1 (resp. 0), if the semantic of $\hat{d}^{(i)}$ and $ d^{(i)}$ are the same (resp. otherwise). On the other hand, absolute reward is defined based on the Levenshtein distance of the two sequences (denoted by $z(\cdot, \cdot)$): i.e., $r_{\text{abs}}(\hat{d}^{(i)}; d^{(i)};\texttt{LM}_{\boldsymbol{\theta}_{\text{exp}}})$ is defined as:
\begin{align}
    r_{\text{abs}}(\cdot) = \begin{cases}
         1 - \frac{z( \hat{d}^{(i)}, {d}^{(i)})}{\max \{|\hat{d}^{(i)}|, |{d}^{(i)}|\}} & \text{if } z( \hat{d}^{(i)}, {d}^{(i)}) \leq z_0,\\
         0 & \text{otherwise},
    \end{cases}
\end{align}
where $z_0$ is a similarity threshold. By incorporating the above LTI process to on-policy distillation, our on-policy knowledge seeding (KS) objective is defined as:
\begin{align}\nonumber
\mathcal{L}_{\text{KS}}(\boldsymbol{\theta}, \boldsymbol{\theta}_{\text{exp}})
&= \E_{x\sim \mathcal{D}} \Big[
(1-\alpha)
\E_{y \sim \texttt{LM}_{\boldsymbol{\theta}_{\text{exp}}}(\cdot|x)}
\left[r(y)\right]  - \alpha
\E_{y \sim \texttt{LM}_{\boldsymbol{\theta}_{\text{exp}}}(\cdot|x)}
\big[
\D(\texttt{LM}_{\boldsymbol{\theta}} \|
\texttt{LM}_{\boldsymbol{\theta}_{\text{exp}}})(y|x)
\big]
\Big].
\end{align}
where $\alpha \in [0, 1]$ controls the strength of the distillation compared to the LTI objective. Based on this objective, we update the new expanded parameters of the model and consolidate the memory/knowledge of high frequency memory into lower-frequency memory blocks. Now that the memories in $\texttt{MLP}^{(f_{\ell^* - 1})}(\cdot)$ are consolidated in $\texttt{MLP}^{(f_{\ell^*})}(\cdot)$, we reset all the low-rank parameters that previously (in past sleep periods) have been added to $\texttt{MLP}^{(f_{\ell^* - 1})}(\cdot)$, making its capacity available for future. This step, can be interpreted as a similar procedure of synaptic pruning in human brain, in which brain prunes connections that are unnecessarily and/or redundant~\citep{li2017rem} to enhance its efficiency and performance.

\head{Note on the Implementation} Implementing the growing sparse modules can be extremely challenging if it requires a direct change in the dimensionality of tensors in the implementation. Alternatively, we can initially have those parameters in the model, but masked them in the forward and backward pass, before their initial activation in a sleep stage. Interestingly, it also aligns with our understanding of human brain, where brain has (large but) fixed capacity and new components are not added over time. Instead, new connections between brain regions can shape through our life, unlocking the activation of new neurons and resulting in more plasticity to learn new tasks~\citep{kandell2021principles}.

\subsection{Dreaming: A Self-Modifying Process}  \label{sec:dream}
The previous stage, which involved freezing higher-frequency parameters and distilling their knowledge to lower-frequency memories acts similar to slow-wave stage of sleep (NREM) in humans, which is responsible for memory consolidation. In REM stage, however, the brain is highly active (even on par with waking time) and aims to self-modify and strengthen newly formed synapses by dreaming. Inspired by this, we aim to design a dreaming process that learns how to generate dreams (synthetic data) that can help itself to improve over time.

In practice, any synthetic data generation process for self-improvement (e.g., \citet{pang2024language, huang2025selfimprovement, self-adapting-llms}) can be incorporated in this stage. A critical consideration, however, is the risk of iteratively applying self-improvement in continual learning setup, which might cause catastrophic forgetting~\citep{self-adapting-llms}. In our evaluation, we show that how our two-step design of sleep as memory consolidation and then dreaming as self-modifying process is more robust to catastrophic forgetting. As a proof of concept, we build upon the work of \citet{self-adapting-llms}, SEAL; however, there are three challenges to incorporate it in our sleep paradigm: (1) Due to the cost of supervised fine-tuning (SFT) in SEAL's inner-loop, it is limited to small number of self-edits (dreams in our terminology). (2) Potential catastrophic forgetting as the cause of iterative self-improvement in sleep periods. 
(3) The sampling process only samples from the existing knowledge space of the model, while one of the key roles of dreaming is to explore novel synthesis of memories~\citep{stickgold2005sleep}.

Given a sampled task $(C, \tau)$, where $C$ is the context containing information relevant to the task and $\tau(\cdot)$ is a measure to asses the performance in the downstream evaluation, our ``dreaming'' process starts with generating $m \geq 1$ dreams with having $C$ in context. In the sampling process, each router in MoE blocks additionally chooses a random expert and so incorporates random irrelevant knowledge to the dreaming, learning the underlying patterns that are hidden from model's sight. For this step, we let $\{\texttt{DREAM}^{(i)}\}_{i=1}^{m} \sim \texttt{LM}_{\boldsymbol{\theta}}(\cdot | C)$. Next, we reject some of the generated dreams and only keeps the samples with the most potential in improving the model's performance. To this end, we take inspiration from the literature on gradient-based data selection~\citep{wang2024greats, pan-etal-2024-g}: for each dream, $\texttt{DREAM}^{(i)}$, we assign an importance score $\boldsymbol{\omega}^{(i)}$ and select \texttt{Top}-$k$ dreams with highest importance score along with $b$ random samples to maintain diversity. Given language modeling objective $\mathcal{L}_{SFT}(\cdot)$, we define importance score of $\texttt{DREAM}^{(i)}$, denoted as $g^{(i)}_{\texttt{DR}}$, as the gradient of the objective: $g^{(i)}_{\texttt{DR}} = \nabla_{\boldsymbol{\theta}} \mathcal{L}_{SFT}(\texttt{DREAM}^{(i)}, \boldsymbol{\theta}).$ We let $\texttt{D}$ be the set of all selected dreams by the above process. For each $\texttt{DREAM}^{(i)} \in \texttt{D}$ we consider an isolated instance of the model and updates its parameters via supervised
finetuning (with LoRA~\citep{hu2022lora}): i.e., $\boldsymbol{\theta}'^{(i)} \leftarrow \texttt{SFT}\left( \boldsymbol{\theta}^{(i)}, \texttt{DREAM}^{(i)} \right)$. Given the new fine-tuned model, following SEAL~\citep{self-adapting-llms}, we reward the generation of $\texttt{DREAM}^{(i)}$ based on $\texttt{LM}_{\boldsymbol{\theta}'^{(i)}}$'s performance improvement~over~$\texttt{LM}_{\boldsymbol{\theta}^{(i)}}$:
 \begin{equation}
\label{eqn:reward}
r\left(\texttt{DREAM}^{(i)}, \tau(\cdot), \texttt{LM}_{\boldsymbol{\theta}^{(i)}}\right) =
\begin{cases}
1 & \text{If improves},  \\
0 & \text{Otherwise}.
\end{cases}
\end{equation}
We follow SEAL and use ReST$^\text{\textit{EM}}$ algorithm~\citep{singh2024beyond} to optimize the above process.

\begin{figure*}
    \centering
    \includegraphics[width=0.31\linewidth]{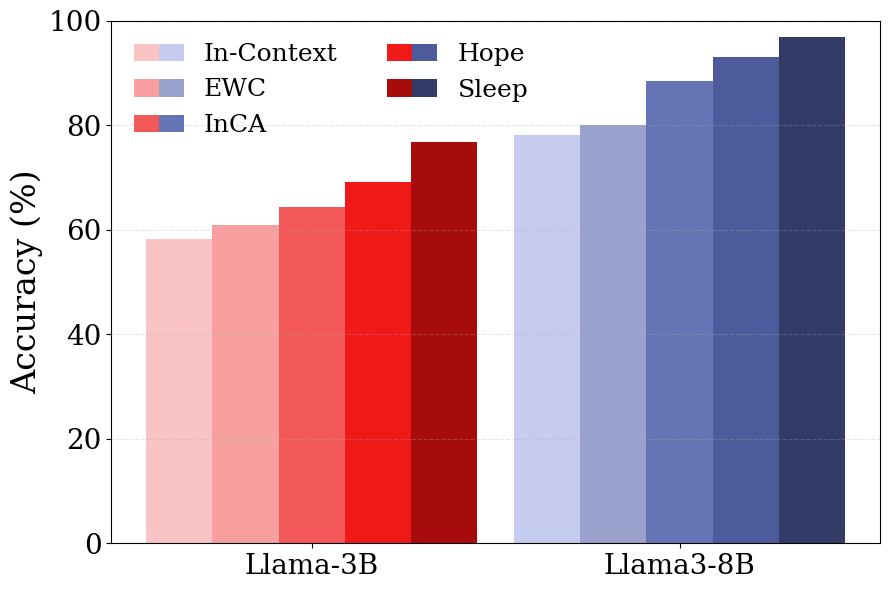}~\hfill~
    \includegraphics[width=0.31\linewidth]{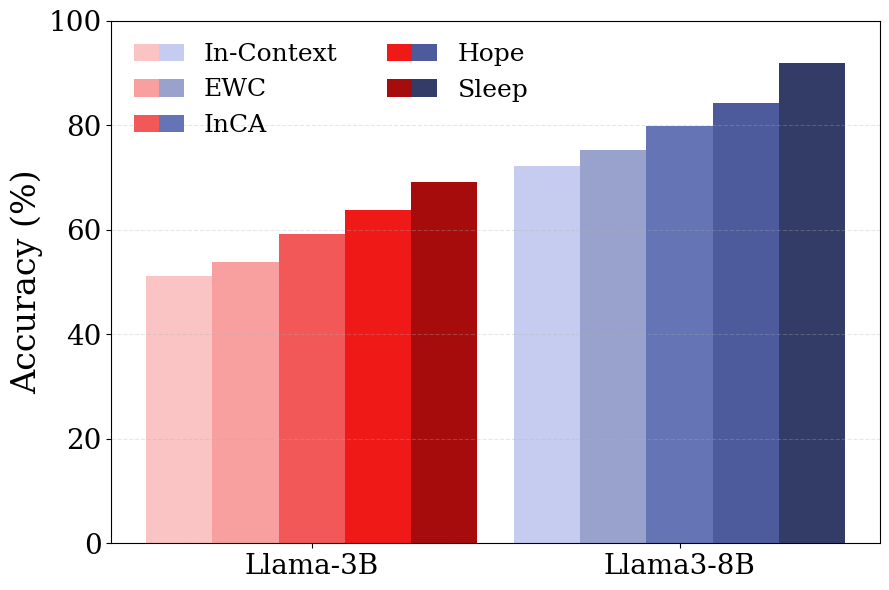}~\hfill~
    \includegraphics[width=0.31\linewidth]{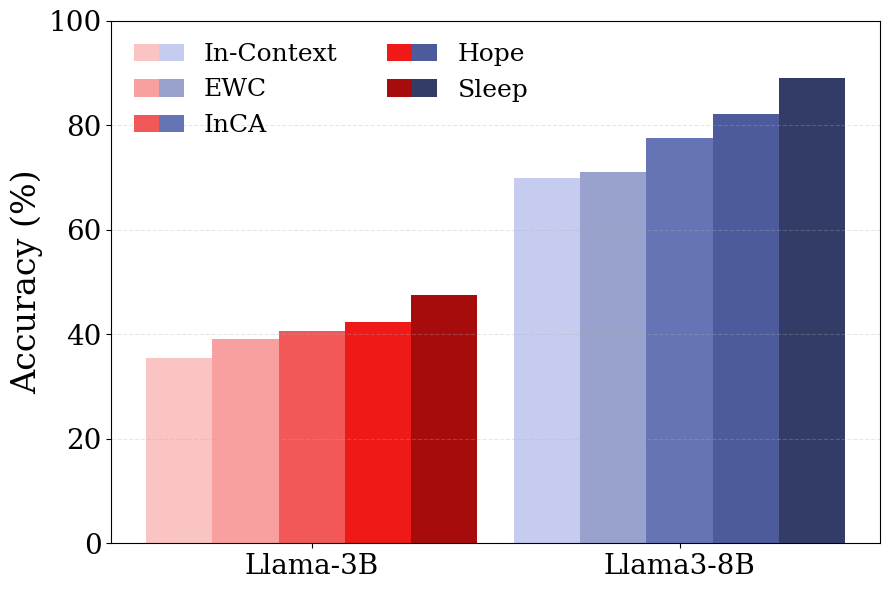}
    \caption{Class-incremental learning for text classification is evaluated on the (Left) CLINC dataset~\citep{larson2019evaluation}, (Middle) Banking dataset~\citep{casanueva2020efficient}, and (Right) DBpedia dataset~\citep{auer2007dbpedia}. The  \model{} architecture consistently outperforms other continual learning approaches, achieving the highest accuracy.}
    \label{fig:CIL}
\end{figure*}

\section{Empirical Results}\label{sec:exp}
In our empirical evaluation, we study the effect of each stage of the sleep and also all the stages together. In the first section, we focus on the memory consolidation, which is designed with the goal of enhancing the continual learning abilitites and long-context understanding.  See \autoref{app:add-exp} for additional experimental results.

\begin{figure*}
    \centering
    \includegraphics[width=0.33\linewidth]{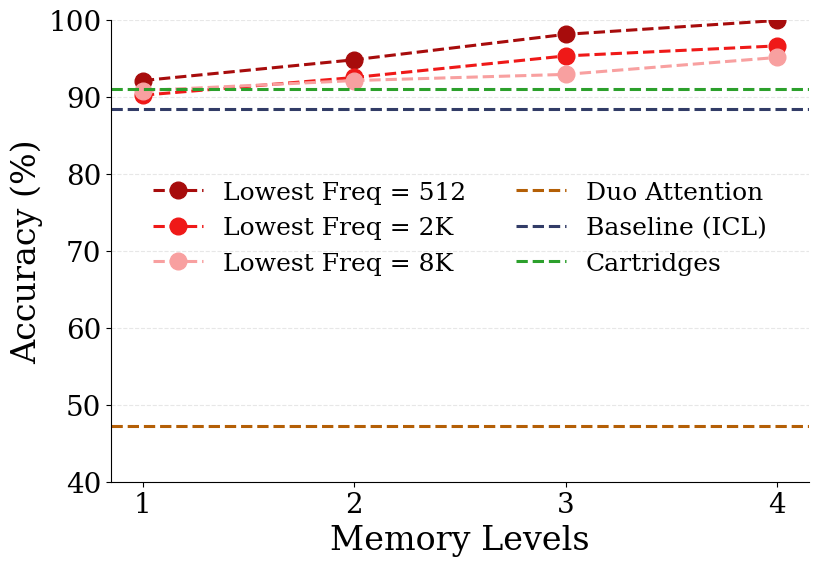}~
    \includegraphics[width=0.33\linewidth]{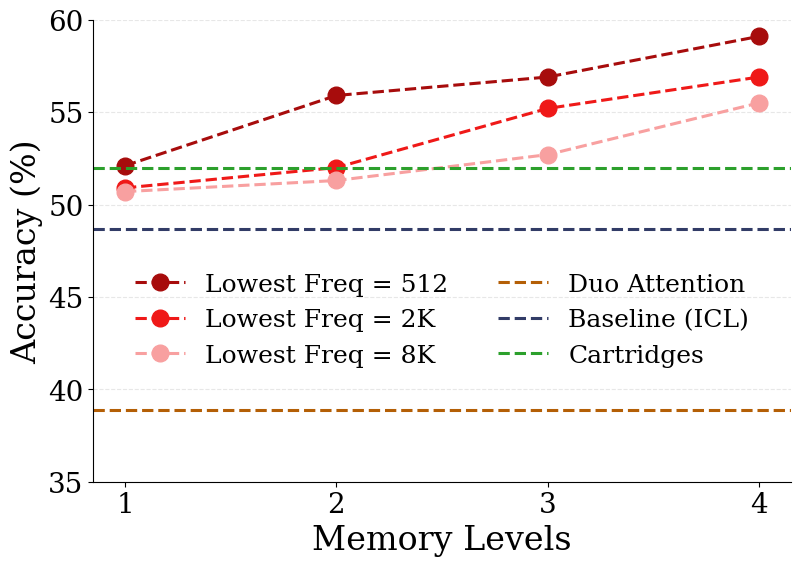}~
    \includegraphics[width=0.33\linewidth]{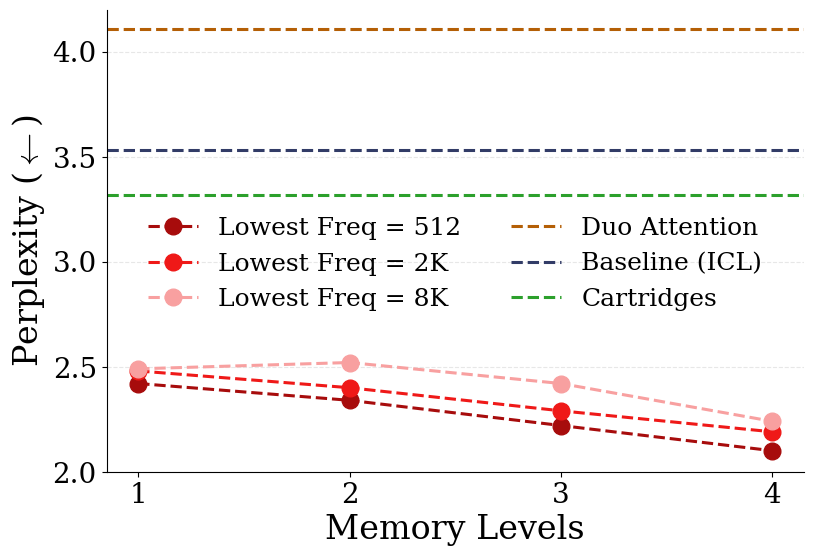}
    \caption{Effect of memory levels on in-context learning performance for (Left) MK-NIAH from RULER~\citep{hsieh2024ruler}, (Middle) LongHealth~\citep{adams2025longhealth}, and (Right) QASPER~\citep{dasigi2021dataset}. Lower values indicate better performance for QASPER.}
    \label{fig:effect-level}
\end{figure*}

\subsection{The Effect of Memory Consolidation}
One of the main goals of memory consolidation process is to improve continual learning while strengthening long-context understanding. In this section, we evaluate \emph{only} memory consolidation phase (without self-improvement) on continual-learning and long-context benchmarks.

\head{Class Incremental Learning}
We first focus on class-incremental learning on three datasets of CLINC~\citep{larson2019evaluation}, Banking~\citep{casanueva2020efficient}, and DBpedia~\citep{auer2007dbpedia} (see \autoref{app:data} for the details). We use Llama-3B and Llama3-8B~\citep{dubey2024llama} as the backbones. \model{} is augmented with our memory consolidation mechanism, which  improves abstraction via on-policy self-distillation. Following \citet{momeni2025context}  we compare against ICL (same continual pre-training process but without sleep), Elastic Weight Consolidation (EWC)~\citep{kirkpatrick2017overcoming}, and In-context Continual Learning with an External Learner (InCA)~\citep{momeni2025context}. We also include Hope ~\citep{behrouz2025nested} as a multi-level in-context updating baseline without explicit distillation process. Results in \autoref{fig:CIL} show that \model{} performs best across datasets, including over external-learner (InCA) and regularization (EWC) approaches. Relative to ICL, gains come from converting prompt-level adaptation into durable parametric memory through consolidation. Relative to Hope, explicit self-distillation yields better abstractions than repeated in-context updates alone. 

\head{The Effect of Levels (\#Sleep Phases) on In-context Learning}
To better isolate how \model{}’s consolidation schedule impacts in-context learning and long-context understanding, we evaluate question answering and multi-key retrieval under long contexts. We use LongHealth~\citep{adams2025longhealth}, QASPER~\citep{dasigi2021dataset}, and MK-NIAH~\citep{hsieh2024ruler} datasets (see \autoref{app:data}). As baselines, we use ICL,   DuoAttention~\citep{xiao2025duoattention}, and Cartridges~\citep{eyuboglu2025cartridges}. For \model{} variants, we vary the sleep schedule by changing (i) how many consolidation stages are used and (ii) the persistence of the most stable memory, operationalized by the lowest consolidation frequency. Intuitively, a lower frequency yields more persistent but less adaptive long-term memory. Results are reported in \autoref{fig:effect-level}. Across all tasks, \model{} consistently outperforms ICL and the efficient DuoAttention baseline, showing that sleep-time consolidation improves long-context behavior beyond prompt-only adaptation. We also observe that \model{} outperforms Cartridges~\citep{eyuboglu2025cartridges}: while Cartridges improves efficiency by using an auxiliary model to compress KV representations, \model{} instead performs on-policy self-distillation during sleep, consolidating newly acquired information into transferable parametric knowledge and yielding more robust long-context understanding. Comparing \model{} variants with each other, we find two consistent trends: (1) increasing the number of consolidation stages improves in-context learning and long-context understanding, supporting the view that sleep enables better knowledge abstraction and compression, allowing the model to retain more information with fewer effective parameters; and (2) increasing the lowest frequency reduces performance, suggesting that making the most persistent memory more adaptive weakens retention.

\begin{figure}[ht]
    \centering
    \begin{minipage}{0.3\linewidth}
        \centering
        \includegraphics[width=1.0\linewidth]{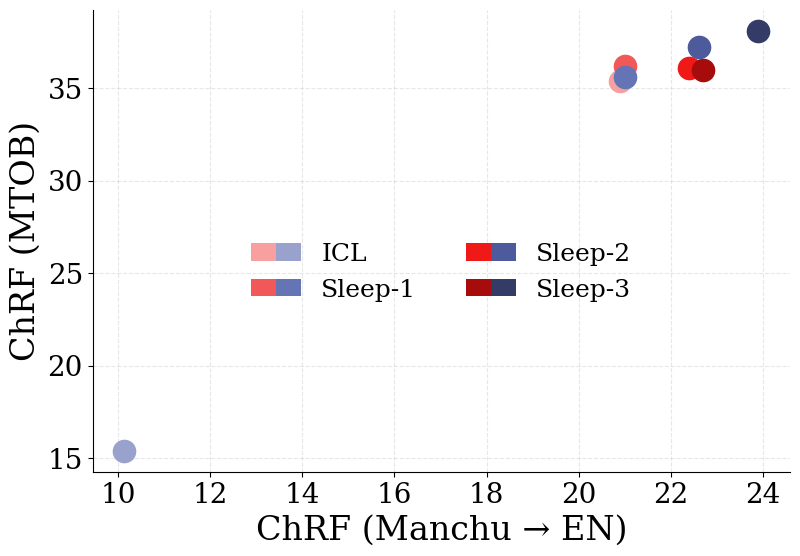}
        \caption{Continual Translation of a Novel Language (CTNL) task. Red points show performance when training on a single language, whereas blue points show performance under continual learning.}
        \label{fig:icl-translate}
    \end{minipage}\hfill
    \begin{minipage}{0.3\linewidth}
        \centering
        \includegraphics[width=1.0\linewidth]{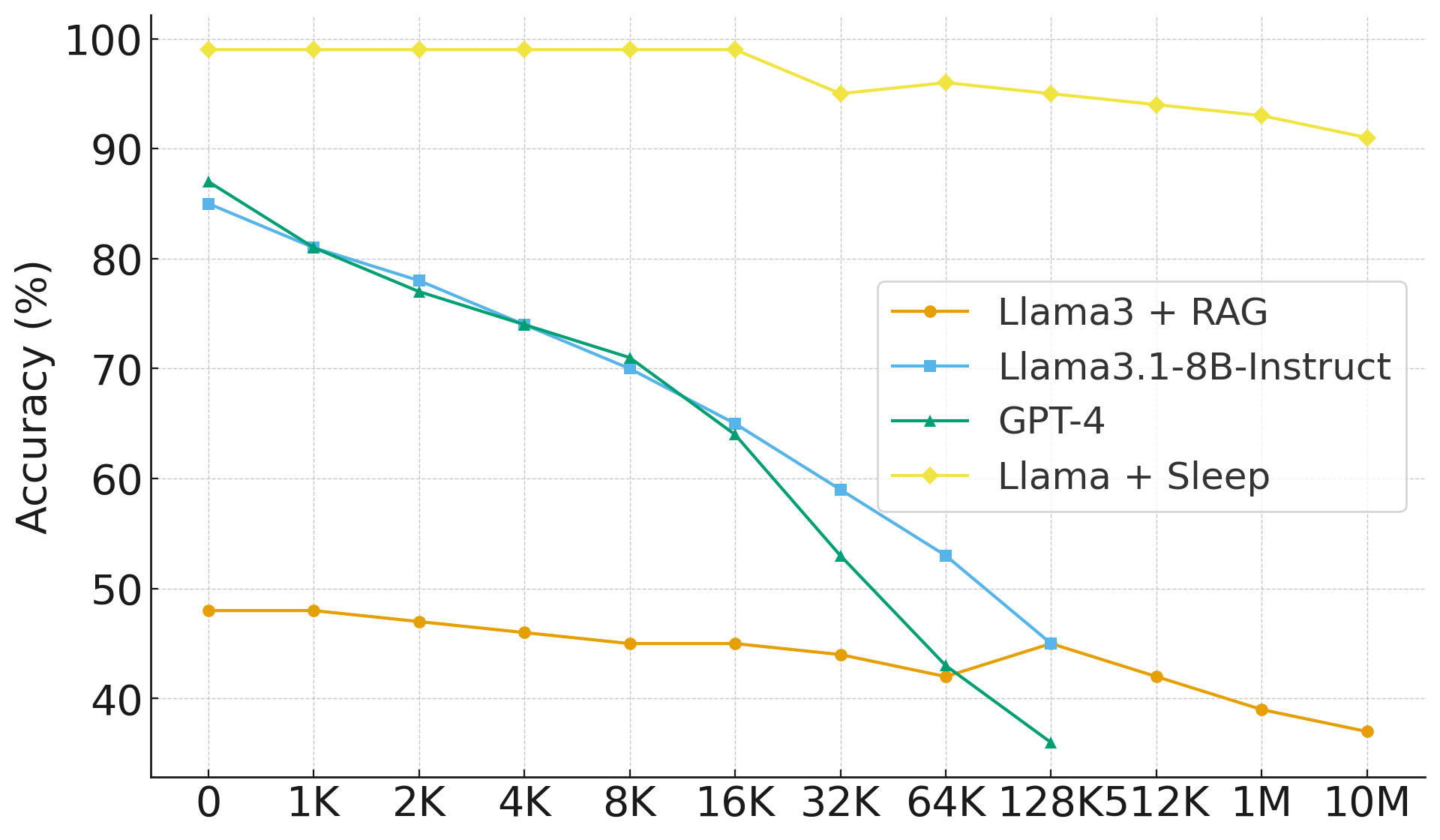}
        \caption{Results on the BABILong benchmark. Red points correspond to fine-tuned models, whereas blue points correspond to zero-shot evaluations of large-scale models.}
        \label{fig:babilong}
    \end{minipage}
    \hfill
    \begin{minipage}{0.36\linewidth}
        \centering
        \captionof{table}{Performance of different methods on mathematical reasoning benchmarks. We use different variants of Qwen models and report average@16.}
        \vspace{1mm}
        \label{tab:ablation}
        \resizebox{\linewidth}{!}{
        \begin{tabular}{lccc}
        \toprule
        \textbf{Method} & \textbf{AIME-24} & \textbf{AIME-25} & \textbf{HMMT-25} \\
        \midrule
        \midrule
        \multicolumn{4}{c}{\textit{Qwen3-8B}} \\
        \midrule
        \rowcolor{mygray}
        Sleep          & 79.2 & 69.0 & 46.1 \\
        - Imitation Learning & 76.8 & 67.9 & 45.0\\
        - Semantic Reward & 78.9 & 69.2 & 44.5 \\
        - w/o Expansion & 78.2 & 67.9 & 44.9 \\
        \midrule
         OPSD           & 76.6 & 67.4 & 45.1\\
         + Expansion    & 77.9 & 68.2  & 45.9\\
        \bottomrule
        \end{tabular}
        }
    \end{minipage}
\end{figure}

\head{Learning a New Language In-Context}
LLMs often fail in continual settings, where models are expected to acquire new skills sequentially without overwriting previously acquired knowledge. To study this challenge, we follow the task introduced by \citet{behrouz2025nested}, which combines MTOB \citep{tanzer2024a} and Manchu \citep{pei2025understanding} datasets, two translation datasets for unseen languages during pre-training. That is, models are exposed in-context to two previously unseen languages and must translate phrases into English. We consider two setups: learning and evaluating each language independently, and sequentially learning both languages before evaluating translation performance on each.
We compare standard ICL with variants of \model{} that differ in the number of sleep-time consolidation stages, denoted as \model-1, \model-2, and \model-3.
Results are shown in \autoref{fig:icl-translate}, where each point reports ChRF scores for Manchu$\rightarrow$English (x-axis) and Kalamang$\rightarrow$English (y-axis). In the single-language setting, all \model{} variants match or exceed ICL, indicating that consolidation does not hinder in-context adaptation. Under continual learning setup, ICL exhibits a sharp performance drop, largely reverting to pre-trained behavior, whereas \model{} retains substantially more of its gains. Performance improves monotonically with additional consolidation stages, and \model-3 nearly recovers its single-language performance despite sequential exposure.
These results underscore the role of sleep-time consolidation in continual learning. Unlike ICL’s prompt-level updates, Sleep introduces an explicit self-improvement phase that distills useful abstractions into longer-lasting parametric memory, enabling effective sequential learning without catastrophic forgetting. As another baseline, we also evaluated Cartridges~\citep{eyuboglu2025cartridges} and Supervised Fine-Tuning (SFT) over the languages. Surprisingly, both methods faced catastrophic forgetting in at least on of the languages, performing even weaker than ICL in at least one of the tasks (placing them outside of the plot in \autoref{fig:icl-translate}).  

\head{BABILong}
We evaluate \model{} (Sleep) on the BABILong benchmark~\citep{kuratov2024babilong}, comparing against:
(1) large models (GPT-4 and GPT-4o-mini~\citep{achiam2023gpt});
(2) a mid-scale Llama-8B model~\citep{dubey2024llama} with RAG; and
(3) state-of-the-art small long-context models, including RMT~\citep{bulatov2022recurrent}, ARMT~\citep{rodkin2024associative}, and Titans~\citep{behrouz2024titans}. Detailed discussion on the results can be found in \autoref{app:BABILong}. In summary, \model{} achieve almost perfect score in scaling to 10M of tokens.

\head{Memory Consolidation for Reasoning}
Another important implication of memory consolidation phase is improving the reasoning capability of the model. In this section, we evaluate its effect of mathematical reasoning and compare it with common baselines of base model, SFT, and GRPO~\citep{shao2024deepseekmath}. The results are reported in \autoref{tab:main_results}. Our algorithm for memory consolidation performs better than SFT and GRPO for improving the reasoning capabilities of the base model. 

\begin{minipage}{0.36\textwidth}
    \centering
\captionof{table}{Performance of different methods on mathematical reasoning benchmarks. We use different variants of Qwen models and report average@16.}
\vspace{1mm}
\label{tab:main_results}
\resizebox{\linewidth}{!}{
\begin{tabular}{lccc}
\toprule
\textbf{Method} & \textbf{AIME-24} & \textbf{AIME-25} & \textbf{HMMT-25} \\
\midrule
\midrule
\multicolumn{4}{c}{\textit{Qwen3-1.7B}} \\
\midrule
 Base (Instruct) & 49.8 & 34.5 & 25.7 \\
 SFT            & 47.3 & 36.1 & 22.9 \\
 GRPO           & 51.0 & 38.6 & 26.1 \\
OPSD           & 51.6 & 40.0 & 28.1\\
\rowcolor{mygray}
Sleep          & 53.2 & 40.2 & 29.3 \\
\midrule
\multicolumn{4}{c}{\textit{Qwen3-8B}} \\
\midrule
Base (Instruct) & 73.8 & 68.1 & 42.4 \\
 SFT            & 75.5 & 66.4 & 43.7 \\
 GRPO           & 76.4 & 68.1 & 44.9 \\
 OPSD           & 76.6 & 67.4 & 45.1\\
\rowcolor{mygray}
Sleep          & 79.2 & 69.0 & 46.1 \\
\bottomrule
\end{tabular}
}
\end{minipage}
\hfill
\begin{minipage}{0.56\textwidth}
    \begin{minipage}{\linewidth}
        \centering
    \captionof{table}{Knowledge Incorporation Performance across Passage Settings}
    \resizebox{\linewidth}{!}{
    \label{tab:merged_knowledge_incorp}
    \begin{tabular}{l c c}
        \toprule
        \multirow{2}{*}{\textbf{Method}} & \textbf{Single Passage } & \textbf{Continued Pretraining } \\ 
        &  (n = 1) & (n = 200)\\
        \midrule
        Base model & 31.9 & 31.9 \\
        Fine-tuned Model with No Dreaming & 33.4 & 32.0 \\
        SEAL & 46.7 & 43.2 \\
        \rowcolor{mygray}
        Sleep (Transformer) & 48.1 & 44.3 \\
        \rowcolor{mygray}
        Sleep (Transformer + four-level) & 48.9 & 46.2 \\
        \midrule
        removing gradient-based selection &	47.1 &	45.2 \\
        removing random expert &	48.0 &	44.7 \\
        removing Dreaming & 35.7 &	36.2 \\
        \bottomrule
    \end{tabular}
    }
    \end{minipage}
    \begin{minipage}{0.9\linewidth}
        \centering
        \vspace{2ex}
    \captionof{table}{Few-shot Abstract Reasoning}
    \resizebox{0.5\linewidth}{!}{
    \begin{tabular}{l c}
        \toprule
        \textbf{Method} & \textbf{Success Rate (\%)} \\
        \midrule
        ICL & 0 \\
        TTT & 10 \\
        SEAL & 72.5 \\
        \rowcolor{mygray} Sleep  & 80 \\
        \bottomrule
    \end{tabular}
    }
    \label{tab:few-shot-methods_comparison}
    \end{minipage}
\end{minipage}

\head{Knowledge Incorporation} 
Next, we focus on the full design of Sleep, allowing for self-improvement as well. In this task, we expect the model to be able to answer questions about the incorporated facts.  We follow the experimental setup of \citet{self-adapting-llms}, including the choice of models and parameters, for the sake of fair comparison. We evaluate our model on integrating new factual information from \SQuAD dataset~\citep{rajpurkar2016squad}. As baselines, we use (i) a base model, which is the variant without any improvement or having access to the passage; (ii) a fine-tuned model with no dreaming, (iii) SEAL model with RL and self-adaption; (iv) our Transformer-based architecture with two level memory system; and (v)  with four-level memory system. 

Table~\ref{tab:merged_knowledge_incorp} summarizes mean no-context SQuAD accuracy for both the single-passage ($n=1$) and continued pretraining (CPT, $n=200$) settings. Our sleep process achieves the best results among other settings and state-of-the-art methods like SEAL. We attributes this results to: (1) memory consolidation steps that let the model store its knowledge more effectively; (2) our improvements on top of the SEAL that we discussed in \autoref{sec:dream}. In the CPT regime, the model is exposed to $n=200$ passages during a single continued pretraining run and is evaluated on the full set of 974 associated questions. For each passage, we sample five dreams and combine them into an aggregated synthetic dataset for training.  Sleep process achieves the best performance.

\head{Few-Shot Learning} \label{sec:few-shot}
We  follow the few-shot ARC experimental protocol from prior work~\citep{akyureksurprising, self-adapting-llms} and adapting it to our \textsc{Sleep} paradigm. As the backbone we use \texttt{Llama-3.2-1B}. Following common practice, we filter  subset of data to avoid tasks that remain unsolvable under standard configurations, yielding 11 tasks for training and 8 held-out tasks for evaluation. See \autoref{app:arc} for the details. In this setting (see \autoref{tab:few-shot-methods_comparison}), \textsc{Sleep} achieves a 80\% success rate, higher then the other methods.

\vspace{-1ex}
\subsection{Ablations on the Design Choices}
We ablate the design choices for the memory consolidation process on the mathematical reasoning. The results are reported in \autoref{tab:ablation}. All the components contribute positively to the performance of our method. We also, further ablate the design choices on self-improvement in \autoref{tab:merged_knowledge_incorp}.

\section{Conclusion}
In this work, we introduced the \textsc{Sleep} paradigm for Large Language Models, consists of: (i) \emph{knowledge seeding}, an upward distillation that transfers short-term, in-context knowledge into lower-frequency, long-term parameters, and (ii) \emph{dreaming}, self-generated training that improves capabilities while controlling interference. In our experimental results, across long-context understanding, knowledge incorporation, few-shot reasoning, and continual learning, \textsc{Sleep} yields consistent gains.

\newpage
\printbibliography
\newpage
\appendix

\newpage
\appendix

\begin{figure}[ht]
    \centering
    \includegraphics[width=1\linewidth]{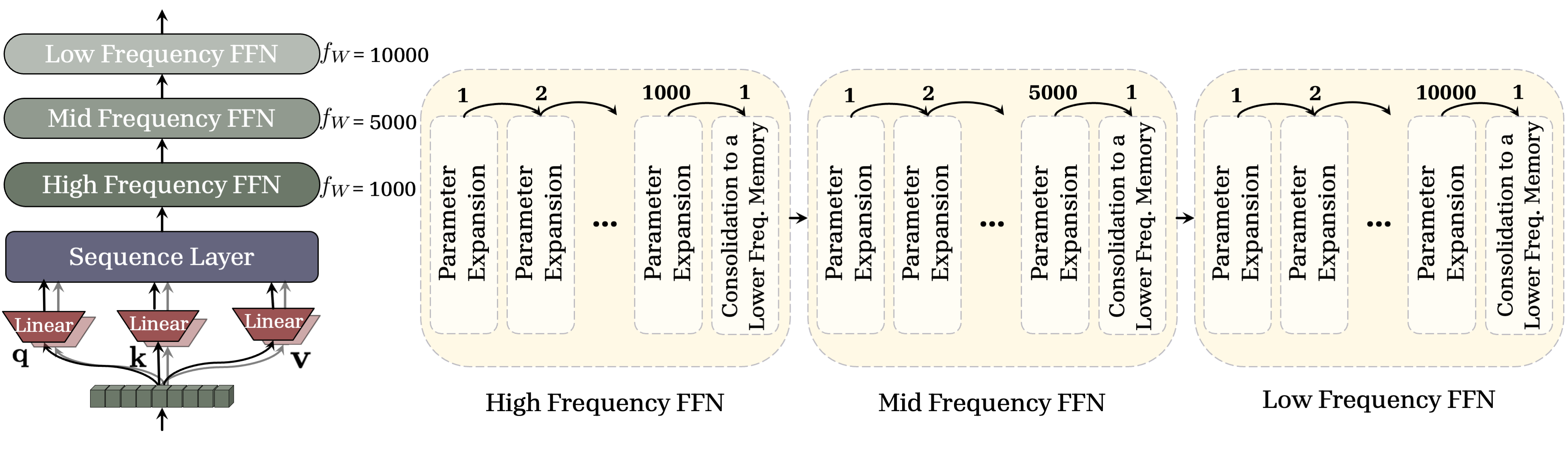}
\caption{Multi-frequency memory hierarchy. Updates enter the High-Frequency FFN via repeated Parameter Expansion; when the window $f_W$ expires, knowledge is Consolidated to the Mid- and then Low-Frequency FFNs (1k→5k→10k).}

    \label{fig:selected-params}
\end{figure}

\begin{figure}[ht]
    \centering
    \includegraphics[width=1\linewidth]{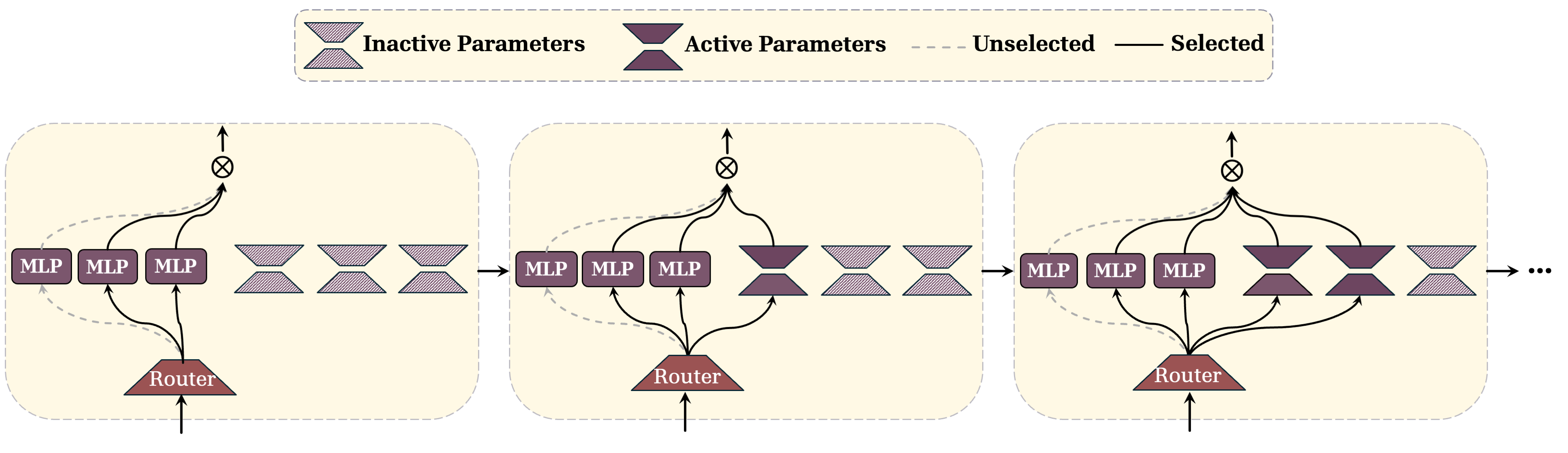}
\caption{Memory consolidation by routed expert updates. Across Sleep cycles (left$\rightarrow$right), a router selects and updates a small set of experts (solid), leaving others inactive (hatched), expanding capacity while limiting interference.}

    \label{fig:frequency}
\end{figure}

\section{Related Work}
\label{app:rw}

\subsection{Parameter-Efficient Adaptation and Composition}
Parameter-efficient fine-tuning (PEFT) adapts large language models (LLMs) to specific tasks by optimizing a minimal set of auxiliary parameters while freezing the backbone. 
\textbf{Low-Rank and Prefix Adaptation.} Prominent methods include LoRA \citep{hu2022lora}, which injects trainable low-rank matrices into linear projections, and prefix/prompt-tuning \citep{li2021prefix,lester2021power}, which steers computation via learnable virtual tokens. Recent variants optimize initialization to accelerate convergence, such as PiSSA \citep{meng2024pissa}, or extend these mechanisms for context-to-parameter distillation \citep{eyuboglu2025cartridges}.

\textbf{Composition and Routing.} Beyond single-task adaptation, research increasingly focuses on composing multiple adapters. Approaches include weighted composition via in-context learning \citep{huang2023lorahub} and retrieval-based routing, where relevant LoRA modules are selected dynamically per input \citep{zhao2024loraretriever}. Advanced configurations utilize mixture-of-experts (MoE) architectures to activate specific low-rank paths or merge adapter parameters dependent on the task \citep{xiao2024configurable,yadav2024survey,wu2024mixture,gou2023mixture,li2024mixlora,zhao2024merging}.

\subsection{Knowledge Injection, Distillation, and Self-Improvement}
\textbf{Parametric Knowledge Injection.} 
To reduce reliance on retrieval at inference time, recent work injects external knowledge directly into model parameters. Techniques range from \emph{Parametric RAG}, which assigns document-specific LoRA adapters \citep{su2025parametricrag}, to prompt distillation, where student models learn from teacher-generated QA pairs or synthetic conversations \citep{kujanpaa2024knowledge,mao2025lift,caccia2025training}. \emph{Cartridges} \citep{eyuboglu2025cartridges} bridge PEFT and distillation by pre-training a reusable "KV cache adapter" via a self-study objective, achieving in-context learning (ICL) quality with reduced serving costs.

\textbf{Self-Improvement and Meta-Learning.} 
Complementary to distillation is the use of model-generated signals for self-improvement. This includes Reinforcement Learning (RL) on verifiable rewards to enhance reasoning \citep{deepseekai2025r1,singh2023beyond,zelikman2022starbootstrappingreasoningreasoning} and self-rewarding mechanisms \citep{wang2025cream,pang2024selfimprove}. Meta-learning frameworks, such as SEAL \citep{self-adapting-llms}, extend this by optimizing the adaptation strategy itself—learning \emph{how} to generate self-edits—drawing on broader principles of self-referential learning \citep{schmidhuber1987meta,finn2017maml,irie2022modern}. These methods rely heavily on high-quality synthetic data generation to scale supervision \citep{nayak2024learning,abdin2024phi,riaz2025metasynth}.

A recent group of concurrent studies has suggested to use on-policy distillation for self-distillation process~\citep{zhao2026self, hubotter2026reinforcement, zhang2026opsdl} or for continual learning~\citep{shenfeld2026self}. In addition to the fact that this study is older than such methods, Sleep fundamentally delivers a different messages that: (1) Sleep uses an \emph{upward distillation of self}, where it unlocks parameters. (2) Sleep is based on Generalized Distillation method that combines on-policy method with an RL method. (3) Sleep suggest a periodic process where knowledge stores in modules in different frequency of update. 

\subsection{Efficient Context Processing}
\label{app:related-work-reducekv}
Addressing the memory bottleneck of long-context processing involves compressing the KV cache or modifying the attention architecture.

\textbf{Prompt and Cache Compression.} 
Approaches to reduce memory footprints fall into two categories: \emph{Prompt compression} shortens inputs via token filtering (hard-token) \citep{li2023unlocking,jiang2023llmlingua} or by learning compact soft-token embeddings via auto-encoders \citep{chevalier2023adapting,ge2023context,tan2024lloco}. \emph{KV cache compression} operates at runtime, employing eviction policies to drop non-essential keys \citep{tang2024quest,zhang2023h2o,li2024snapkv} or merging tokens based on similarity and attention density \citep{wan2024d2o,liu2024minicache,zhang2024cam,wang2024model}. Low-rank projections of the KV cache have also shown promise in maintaining performance at high compression rates \citep{zhang2024lorc,chari2025kv}.

\textbf{Architectural Innovations.} 
Structural changes to Multi-Head Attention (MHA) include Multi-Query and Grouped-Query Attention (MQA/GQA), which share KV heads to reduce memory bandwidth \citep{shazeer2019fast,ainslie2023gqa}, and linearization techniques that remove the softmax to achieve fixed-size states ($K^\top V$) independent of sequence length \citep{arora2024simple,gu2023mamba,beck2024xlstm}. Recent "learning to compress" architectures, such as Titans and TTT, utilize gradient-based updates on constant-sized memory objects to handle infinite contexts \citep{behrouz2024titans,sun2024learning}. Finally, orchestration systems like MemGPT manage context via virtual memory paging rather than architectural modification \citep{packer2023memgpt}. In a similar direction to compression in the token space, \citet{lin2025sleep} present sleep-time compute. Despite similarity in the name, their method is fundamentally different from our work. While this method aim to find a good summary of the interactions of the model with users in the text space, when the model is idle, our proposal is on using distillation to transfer text data knowledge into a form of parametric weight update. 
\subsection{Recent Concurrent and/or Later Work on On-Policy Distillation}
\paragraph{Concurrent and Subsequent Work on On-Policy Self-Distillation.}
Concurrent and subsequent to the initial version of this work, a rapidly
growing line of literature has explored \emph{on-policy self-distillation}
(OPSD), in which the same model plays the role of both teacher and student
under different conditioning contexts and the student is supervised on its
own rollouts via a per-token divergence against the teacher. The canonical instantiation is OPSD~\citep{zhao2026selfdistilledreasoneronpolicyselfdistillation}, which
conditions the teacher on a verified reasoning trace and minimises a
per-token reverse-KL on the student's own rollouts, yielding gains on
mathematical reasoning over both GRPO and off-policy distillation.
A wave of follow-up works varies the form of the privileged context:
On-Policy Context Distillation~\citep{ye2026onpolicycontextdistillationlanguage} conditions the teacher on transient
in-context information (historical solution traces or optimised system
prompts) so that the student internalises that information into its
parameters; GATES~\citep{stein2026gatesselfdistillationprivilegedcontext} uses document-grounded privileged context together
with a consensus-gating mechanism that handles unreliable supervision by
sampling multiple tutor traces and gating learning by their agreement;
SD-Zero~\citep{he2026selfdistillationzeroselfrevisionturns} conditions a ``reviser'' on the student's response and a
binary reward, then distils the reviser back into the generator, converting
sparse outcome rewards into dense token-level supervision;
and COPSD~\citep{liu2026crosslingualonpolicyselfdistillationmultilingual} transfers reasoning behaviour to low-resource languages
by giving the teacher an English translation and reference solution as
privileged crosslingual context. Apple's ``embarrassingly simple''
self-distillation~\citep{zhang2026embarrassinglysimpleselfdistillationimproves} sits at the limit of this spectrum: it drops the
explicit teacher altogether and supervised-finetunes on the model's own
samples under different decoding configurations, showing that even this
minimal recipe meaningfully improves code generation.
 
\paragraph{Continual, Experiential, and Online Adaptation.}
A second cluster of works applies self-distillation to continual,
experiential, or online improvement. SDFT~\citep{shenfeld2026selfdistillationenablescontinuallearning} explicitly targets
continual learning from demonstrations by using a demonstration-conditioned
model as an on-policy teacher, reducing catastrophic forgetting compared
with standard SFT. OEL~\citep{ye2026onlineexperientiallearninglanguage} couples context distillation with deployment:
it extracts transferable experiential knowledge from interaction
trajectories and consolidates it into parameters via on-policy context
distillation, iterating both stages to form a closed online-learning loop.
$\pi$-Play~\citep{zhang2026piplaymultiagentselfplayprivileged} pushes this idea further into a multi-agent self-play
regime, using the question-construction path produced by the task proposer
as privileged context for the teacher and removing the need for external
data or human feedback in training deep search agents. These works are
closely aligned with our motivation that LLMs should not remain static after
deployment. However, they formulate continual or experiential improvement as
a post-training objective on a \emph{fixed-capacity} checkpoint. In contrast,
our Sleep paradigm treats continual learning as a \emph{memory-system}
problem: newly acquired information is first represented in fast, fragile
memory modules and then periodically consolidated into slower, more stable
parameters via Knowledge Seeding, accompanied by gradual capacity expansion
and the resetting of higher-frequency memory after consolidation. 
 
\paragraph{Application-Specific OPSD Recipes.}
A third group of papers adapts the OPSD template to specific deployment
regimes. CRISP/OPSDC~\citep{sang2026crispcompressedreasoningiterative} conditions the teacher on a ``be concise''
instruction prefix and uses per-token reverse-KL on student rollouts to
compress long chain-of-thought traces without entropy collapse.
MTP~\citep{kirchenbauer2026multitokenpredictionselfdistillation} converts a pretrained next-token-prediction model into a
multi-token predictor with a single online self-distillation objective,
yielding more than $3\times$ inference acceleration on GSM8K at $<\!5\%$
accuracy drop. OPSDL~\citep{zhang2026opsdlonpolicyselfdistillationlongcontext} attacks long-context generation by using the
model's own short-context behaviour on extracted relevant spans as a
self-teacher for its long-context student. Skill-SD~\citep{wang2026skillsdskillconditionedselfdistillationmultiturn} targets
multi-turn agents: completed trajectories are summarised into compact
natural-language ``skills'' that condition only the teacher while the
student trains under the plain task prompt. MSD~\citep{qin2026multilingualsafetyalignmentselfdistillation} moves OPSD into
the multilingual setting by transferring safety behaviour from high-resource
to low-resource languages using only multilingual queries. Collectively,
these works show that self-distillation under privileged conditioning is a
general and flexible post-training recipe across reasoning, efficiency,
agentic, and multilingual axes -- but in each case it is deployed as a
single-objective procedure on a fixed model and a fixed application.
 
\paragraph{Limitations of OPSD and Motivation for the Sleep Paradigm.}
Despite the empirical success of OPSD across these domains, recent analyses
have begun to expose its failure modes. \citet{kim2026doesselfdistillationsometimesdegrade} show that, in
mathematical reasoning, conditioning the teacher on rich privileged
information can suppress the model's epistemic verbalisation (its expression
of uncertainty during reasoning), enabling fast in-domain optimisation at
the cost of severe out-of-distribution degradation, with drops of up to
$40\%$ across Qwen3-8B, DeepSeek-Distill-Qwen-7B, and Olmo3-7B-Instruct.
Several recent OPSD works also report that na\"{i}vely iterating
self-distillation in a long-horizon training pipeline can lead to
information leakage from the privileged teacher and to training
collapse~\citep{wang2026skillsdskillconditionedselfdistillationmultiturn,he2026selfdistillationzeroselfrevisionturns}. These results suggest that self-distillation alone is
not sufficient for robust continual improvement: an iterative
self-improvement loop applied directly to a fixed model can erode prior
capabilities or destabilise reasoning. Our two-stage Sleep framework
directly addresses this concern by separating memory consolidation from
self-improvement: consolidation first stabilises newly acquired knowledge in
freshly expanded lower-frequency parameters before Dreaming further modifies
the model, reducing the risk that iterative self-training overwrites useful
prior capabilities.
 
\paragraph{Positioning of Our Work.}
Our work differs from the OPSD literature along four axes that together
motivate the Sleep paradigm.
\textbf{(i) Upward distillation as memory consolidation.} All of the
self-distillation works listed above share the same backbone between teacher
and student and differ only in conditioning context -- privileged
trace~\citep{zhao2026selfdistilledreasoneronpolicyselfdistillation,kim2026doesselfdistillationsometimesdegrade}, demonstration~\citep{shenfeld2026selfdistillationenablescontinuallearning}, document~\citep{stein2026gatesselfdistillationprivilegedcontext}, context or
system prompt~\citep{ye2026onpolicycontextdistillationlanguage,ye2026onlineexperientiallearninglanguage}, ``concise'' prefix~\citep{sang2026crispcompressedreasoningiterative}, decoding
configuration~\citep{zhang2026embarrassinglysimpleselfdistillationimproves}, skill summary~\citep{wang2026skillsdskillconditionedselfdistillationmultiturn}, reviser
context~\citep{he2026selfdistillationzeroselfrevisionturns}, question-construction path~\citep{zhang2026piplaymultiagentselfplayprivileged}, short-context
substring~\citep{zhang2026opsdlonpolicyselfdistillationlongcontext}, or English reference~\citep{qin2026multilingualsafetyalignmentselfdistillation,liu2026crosslingualonpolicyselfdistillationmultilingual}. In contrast, our
Knowledge Seeding step is an \emph{upward} distillation: a smaller,
higher-frequency memory module is distilled into a strictly \emph{larger}
set of newly expanded low-rank experts in a lower-frequency memory module.
This reframes catastrophic forgetting as a problem of insufficient capacity
rather than of sampling distribution, and addresses it by gradual parameter
growth between consolidation steps rather than by re-conditioning a fixed
teacher. \textbf{(ii) Continuum of memory frequencies.} Where the above
works treat distillation as a flat student/teacher pair, our method operates
over a chain of memory blocks with strictly ordered update frequencies and
performs consolidation between each pair of consecutive blocks; to our
knowledge, only SDFT~\citep{shenfeld2026selfdistillationenablescontinuallearning} explicitly targets the continual-learning
desideratum that motivates us, and it does so without architectural growth
or hierarchical memory. \textbf{(iii) Sleep as a two-phase process beyond
consolidation.} The OPSD literature focuses almost entirely on the
consolidation step. Our framework additionally includes a Dreaming phase
analogous to REM sleep, in which the model generates a curriculum of
synthetic data, weights samples by a gradient-based importance score, and
exploits the MoE router to inject controlled novelty -- explicitly designed
to be robust to the iterative self-improvement failure modes flagged
by~\citep{kim2026doesselfdistillationsometimesdegrade,he2026selfdistillationzeroselfrevisionturns}. \textbf{(iv) Knowledge seeding augments on-policy
distillation with imitation learning.} Where the on-policy distillation
objective in works such as~\citep{zhao2026selfdistilledreasoneronpolicyselfdistillation,ye2026onpolicycontextdistillationlanguage,stein2026gatesselfdistillationprivilegedcontext,sang2026crispcompressedreasoningiterative,ye2026onlineexperientiallearninglanguage,zhang2026opsdlonpolicyselfdistillationlongcontext,qin2026multilingualsafetyalignmentselfdistillation,liu2026crosslingualonpolicyselfdistillationmultilingual} reduces to a per-token
reverse-KL on student rollouts, our Knowledge Seeding objective augments
GKD-style on-policy distillation with an RL-based Learning-to-Imitate term
that jointly rewards semantic and Levenshtein-level alignment with the
teacher's sampling distribution, enabling the larger student not only to
inherit the teacher's knowledge but also to imitate the way the teacher
uses it.

\section{Additional Experimental Results and Details} \label{app:add-exp}
In all of our experiments, we follow the settings of the original benchmark (cited in each section). Therefore, for the sake of space and not duplicating information, we refer to those studies for the details of the models. In our design, we use 5 MLP blocks with dimension 64 as the additional parameters and keep the active parameter count unchanged (I.e., the same as the base model, which is either 8B or 3B). 

\subsection{Datasets}\label{app:data}
\head{Class Incremental Learning}
We focus on class-incremental learning on three datasets:
\begin{itemize}
    \item CLINC~\citep{larson2019evaluation}: CLINC150 is a multi-domain intent classification benchmark for task-oriented dialog, and it is commonly used to evaluate both in-scope intent prediction and out-of-scope (OOS) detection. It contains 150 in-scope intents spanning 10 domains, with 23.7K total queries (22.5K in-scope and 1.2K OOS).
    
    \item Banking~\citep{casanueva2020efficient}: Banking77 is a single-domain intent dataset of short banking customer-service queries, labeled with 77 fine-grained intents (e.g., card issues or PIN resets). It includes 13{,}083 examples, and the intent distribution is noticeably imbalanced.
    
    \item DBpedia~\citep{auer2007dbpedia}: DBpedia-based classification maps Wikipedia-derived descriptions (typically abstracts) to ontology categories (e.g., book, film, animal, place). We use the 70-class, level-2 setting and subsample 10K training and 1K test instances for our experiments.
\end{itemize}

\head{The Effect of Levels on In-context Learning}
To better isolate how \model{}’s consolidation schedule impacts in-context learning and long-context understanding, we evaluate question answering and multi-key retrieval under long contexts. We use the following datasets for this evaluation:

\begin{itemize}
    \item LongHealth~\citep{adams2025longhealth}: LongHealth is a long-context clinical multiple-choice QA benchmark built from lengthy fictional patient-case records. It includes 20 case documents (about 5.1K--6.8K words each); we evaluate on 200 questions sampled from these records.
    
    \item QASPER~\citep{dasigi2021dataset}: QASPER is an information-seeking QA benchmark grounded in full-text NLP papers, with roughly 5K questions over about 1.6K papers. We use each paper’s full text as the context.
    
    \item MK-NIAH~\citep{hsieh2024ruler}: MK-NIAH is the multi-key needle-in-a-haystack task from RULER~\citep{hsieh2024ruler}, where multiple key--value facts are embedded in a long context and the model must retrieve the value for a queried key.
\end{itemize}

\subsection{Additional Results: BABILong}\label{app:BABILong}
We evaluate \model{} (Sleep) on the BABILong benchmark~\citep{kuratov2024babilong}, comparing against:
(1) large models (GPT-4 and GPT-4o-mini~\citep{achiam2023gpt});
(2) a mid-scale Llama-8B model~\citep{dubey2024llama} with RAG; and
(3) state-of-the-art small long-context models, including RMT~\citep{bulatov2022recurrent}, ARMT~\citep{rodkin2024associative}, and Titans~\citep{behrouz2024titans}. 
All small models are fine-tuned using the official BABILong training protocol~\citep{kuratov2024babilong}.
As shown in \autoref{fig:babilong}, large models exhibit rapid degradation as context length increases, failing beyond 128K–256K tokens. RAG improves robustness at longer contexts but still degrades as sequence length grows. Among fine-tuned small models, Titans, ARMT, and \model{} perform comparably up to approximately 1M tokens; beyond this point, Titans and ARMT degrade sharply, whereas \model{} remains stable even at extreme lengths up to 10M tokens.
This robustness is driven by Sleep’s explicit consolidation and self-improvement mechanism, which transforms short-lived activations into compact parametric representations. By learning higher-level abstractions during consolidation, \model{} retains task-relevant information more efficiently, reducing sensitivity to increasing context length.
Finally, we observe that all small models, including \model{}, suffer substantial performance drops without fine-tuning. Compressing ultra-long contexts requires sufficient capacity at lower-frequency memory levels to identify and preserve critical information; fine-tuning enables these components to adapt rapidly and coordinate effectively with high-frequency memory during inference.

\subsection{Additional Experiments: ARC}\label{app:arc}
We  follow the few-shot ARC experimental protocol from prior work~\citep{akyureksurprising, self-adapting-llms} and adapting it to our \textsc{Sleep} paradigm. As the backbone we use \texttt{Llama-3.2-1B}. Following common practice, we filter  subset of data to avoid tasks that remain unsolvable under standard configurations, yielding 11 tasks for training and 8 held-out tasks for evaluation. During each \emph{Sleep} cycle, the model first consolidate its previous memories, and then \emph{dreams} by generating synthetic experiences from the few-shot demos. For each task, we sample 60 dreams and reject 45 of them. At test time, for each unseen task, the model generates 5 dreams and applies them independently before predicting the held-out output. We report the fraction of dreams that yield a correct answer. As the baselines, we follow \citet{self-adapting-llms} and use: (i) ICL (In-Context Learning); (ii) TTT + synthetic updates (no dreaming); and (iii) SEAL~\citep{self-adapting-llms}. In this setting, \textsc{Sleep} achieves a 80\% success rate, higher then the other methods.

\subsection{Additional Details}

\begin{table}[htbp]
\centering
\caption{Training Configuration for GRPO, SFT, and Sleep}
\begin{tabular}{lcc|c}
\toprule
\textbf{Parameter} & \textbf{GRPO} & \textbf{SFT} & \textbf{Sleep} \\
\midrule
\midrule
Learning Rate & $5 \times 10^{-6}$ & $5 \times 10^{-6}$ & $5 \times 10^{-6}$ \\
Effective Batch Size & 32 & 32 & 32 \\
Training Steps & 500 & 100 & 100 \\
\midrule
LoRA Rank ($r$) & 64 &  64  & 64 \\
LoRA Alpha ($\alpha$) & 128 &  128 & 128 \\
\bottomrule
\end{tabular}
\end{table}

\subsection{Efficiency}
With the same number of steps, SFT is 4x more efficient than our method, however, their performance is not comparable. Accordingly, we also compare the efficiency, when targeting a specific performance. We trained the models so they achieve the same performance in AIME-24, AIME-25, and HMMT-25. In this case, SFT requires 4.3x, 3.6x, and 4.8x wall-clock time to match the performance of our design. Therefore, from this perspective, Sleep is very efficient, when targeting a specific performance.


\end{document}